\documentclass[12pt]{article}
\usepackage[margin=1in]{geometry}
\usepackage{setspace}

\usepackage{amssymb}
\usepackage{amsmath}
\usepackage{graphicx}
\usepackage{booktabs, siunitx, multirow}
\usepackage[breaklinks,colorlinks]{hyperref}
\usepackage{subcaption}
\usepackage[most]{tcolorbox}
\usepackage[labelfont=bf, labelsep=period]{caption}

\usepackage[utf8]{inputenc}
\usepackage[numbers,compress]{natbib}

\usepackage{titlesec}
\titleformat{\section}
  {\normalfont\large\bfseries}
  {\thesection}{1em}{\MakeUppercase}

\titleformat{\subsection}
  {\normalfont\normalsize\bfseries}
  {\thesubsection}{1em}{}

\titleformat{\paragraph}[runin]
  {\normalfont\normalsize\bfseries}
  {\theparagraph}{1em}{}[. ]

\titlespacing*{\paragraph}
  {0pt}
  {1ex plus .1ex minus .2ex}
  {1em}

\usepackage{xspace}
\newcommand{\myTitle}{Sidewalk Moments: Are Richer Representations Always More Human-Aligned? Evidence from City-Walk Videos}
\newcommand{\dataset}{CityWalk Corpus\xspace}
\newcommand{\cmark}{\checkmark}
\newcommand{\xmark}{\times}

\begin{document}

\begin{center}
  {\large\bfseries \myTitle}\\[1.5em]

  Liu Liu$^{1,2}$, Freya Huying Tan$^{1,2}$, F\'abio Duarte$^{1,3,*}$\\[0.5em]

  {\small
    $^1$ Massachusetts Institute of Technology, 77 Massachusetts Avenue,
    Cambridge, MA 02139, USA\\
    $^2$ City Form Lab, MIT\\
    $^3$ Senseable City Lab, MIT\\[0.3em]
    $^*$Corresponding author:
    \href{mailto:fduarte@mit.edu}{fduarte@mit.edu}
  }
\end{center}

\vspace{1em}

\begin{abstract}
We examine whether richer visual representations yield more human-aligned measures of urban engagement, using 61 first-person city-walk videos from YouTube segmented into over 50,000 ten-second clips and represented across four modalities: spatiotemporal video features, temporally averaged images (TAIs), audio embeddings, and text-based semantic descriptions. Spearman correlation analysis reveals the expected ordering along the temporal-richness continuum, with video features showing the strongest continuous alignment. However, this ordering breaks down under binary classification of high- versus low-engagement moments (the paradigm most commonly used to train perceptual scoring models), where TAIs consistently match or outperform video across most classifiers and quantile thresholds. An independent two-alternative forced-choice study on Amazon Mechanical Turk confirms that this parity reflects human judgment: participants identified engaging moments with comparable accuracy from TAIs and full video clips, while text performed substantially worse and audio remained near chance. Gap analysis reveals a functional dissociation: video features are advantaged in activity-driven scenes with dynamic content, whereas TAIs better align with human judgments in composition-driven scenes dominated by stable spatial structure. These findings challenge the assumption that richer representations are inherently more human-aligned, and suggest that perceptually grounded temporal compression can be a principled alternative to full video encoding.
\end{abstract}

\vspace{0.5em}

\section{Introduction} \label{sec:intro}

Urban perception research has long sought to characterize how people evaluate the built environment, from Lynch's frameworks for legibility and mental mapping~\citep{lynch1960image, lynch_walk_1959} and Appleyard's studies of street conditions~\citep{appleyard1969buildings, appleyard1970styles}, to environmental psychology dimensions such as coherence, complexity, and mystery~\citep{kaplan1982cognition, canter1977psychology, nasar1990evaluative, rapoport2013human}. Contemporary work has scaled these efforts through digital surrogates: online platforms such as Place Pulse aggregate large volumes of pairwise perceptual judgments to support predictive mapping across cities~\citep{salesses2012place, naik2014streetscore, dubey2016deep}, while physiological and mobile sensing approaches capture perceptual responses in real-world settings~\citep{zhang2023assessing, mavros2022mobile, yang2024urban, li2023eye}.

As visual models become increasingly capable, urban perception research has increasingly relied on large collections of street-level imagery and deep visual features to represent how people experience streets and evaluate the built environment. A common computational pipeline has emerged in which researchers collect street images at scale, obtain perceptual annotations through pairwise comparisons \citep{naik2014streetscore}, train predictive models on these labels \citep{zhang2018measuring}, and apply them to infer perceptual qualities across cities \citep{quintana2025global, naik2014streetscore, de2016death, ito2024understanding}. While this approach has enabled unprecedented spatial coverage, it relies almost exclusively on single static images, omitting the temporal dimension of how people actually encounter streets through continuous movement.

Recent advances in spatiotemporal video encoders~\citep{arnab2021vivit, bertasius2021space, tong2022videomae, wang2023videomae} and multimodal foundation models~\citep{girdhar2023imagebind, li2024llava} have substantially expanded the representational options for encoding urban scenes. Yet large-scale perception studies continue to rely predominantly on static images, and it remains unclear whether modern video representations align more closely with human perceptual judgments than simpler, temporally compressed image-based summaries. Our study directly addresses this gap.

This omission raises a natural question: with the increasing availability of egocentric walking videos and advances in spatiotemporal visual models, does preserving richer temporal detail yield representations that better align with human judgment? At first glance, the answer might seem straightforwardly affirmative. Yet evidence from cognitive science suggests otherwise. Human perception does not operate as a continuous stream of raw visual data; instead, observers rapidly extract summary statistics from dynamic scenes, compressing complex sensory inputs into stable impressions shaped by overall spatial composition and a limited number of salient moments \citep{oliva2006building, whitney2018ensemble, manassi2022illusion}. If the human perceptual system systematically smooths or discards fine-grained temporal variation, then computational representations that preserve high-frequency temporal detail may not improve alignment with human judgment, and could in fact distort it. This reasoning motivates the use of temporally averaged images (TAIs) as a deliberate probe. By pixel-wise averaging frames sampled within a short temporal window, TAIs function as a computational analog of the biological smoothing process described by ensemble perception research, filtering transient noise to reveal stable scene structure. A related phenomenon in face perception reinforces this logic: averaged faces are consistently judged as more attractive and representative than any individual face, because compositing amplifies prototypical structure while suppressing idiosyncratic variation \citep{langlois1990attractive}. TAIs apply the same principle to street-level views, foregrounding persistent urban composition over transient elements.

\begin{figure}[htbp]
    \centering
    \includegraphics[width=1\columnwidth]{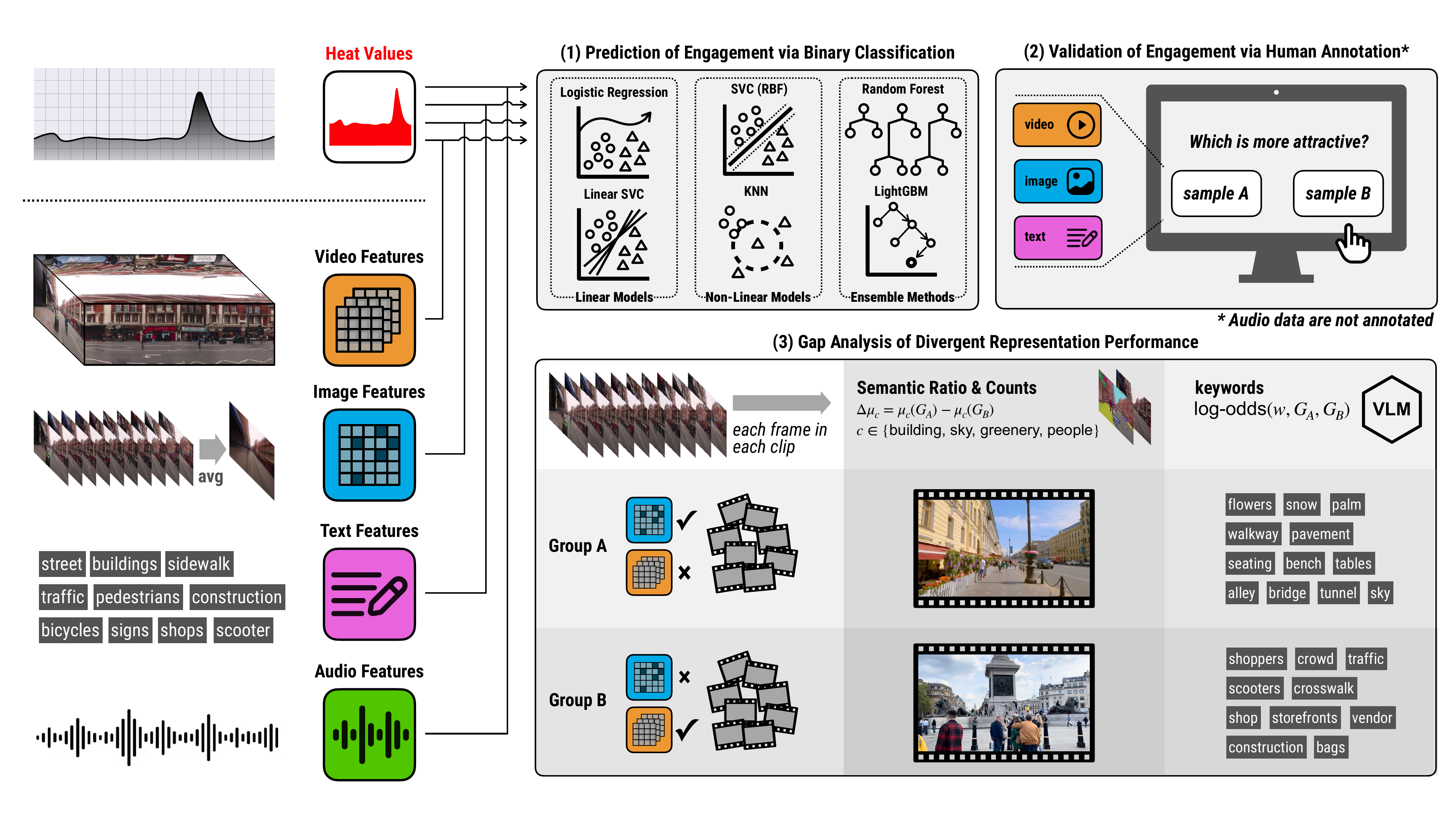}
    \caption{Overview of the study framework. City-walk videos are segmented into short clips and linked to replay-based heat values as a behavioral signal of engagement. Multiple representations are extracted and evaluated through within-video prediction and human preference validation, with a gap analysis examining conditions under which representations diverge.}
    \label{fig:framework}
\end{figure}

To examine this question empirically, we construct a representational comparison framework spanning a temporal richness continuum: from full spatiotemporal video features, through TAIs, to single mid-point frames that carry no temporal information. We operationalize human engagement using replay heat values from 61 egocentric city-walk videos on YouTube, segmenting each recording into over 50,000 ten-second clips represented across four modalities: spatiotemporal video features, TAIs, audio embeddings, and text embeddings from a vision-language model. Representational alignment is evaluated through Spearman correlation analysis, within-video binary classification, an independent Amazon Mechanical Turk validation study, and a diagnostic gap analysis identifying conditions under which representations diverge. Figure~\ref{fig:framework} provides an overview of the study framework. Full details of data collection and feature extraction are provided in the Methods section.

This study makes three contributions. First, it presents a systematic comparison of representations spanning the temporal richness continuum in urban perception, using a consistent feature-extraction backbone to ensure methodological fairness. Second, it demonstrates that TAIs match or outperform video features in binary engagement prediction and that this alignment is confirmed by independent human judgment. Third, it identifies a functional dissociation between representations: video features are advantageous in activity-driven scenes, whereas TAIs better align with human judgments in composition-driven scenes dominated by stable built form. Together, these findings suggest that the choice of representation functions as a methodological filter shaping which dimensions of urban experience become measurable at scale.

\section{Results}\label{sec:results}

Full details of data collection, preprocessing, and feature extraction are provided in the Methods section.

\subsection{Correlation Analysis}\label{ssec:correlation-analysis}

To characterize how well each representation aligns with the engagement signal across the full temporal richness continuum, we computed four ranking-based metrics between clip-level features and replay heat values. Spearman's $\rho$ measures the overall rank correlation between predicted and true engagement orderings. Kendall's $\tau$ provides a concordance-based alternative that is more robust to outliers. Pairwise accuracy captures the proportion of clip pairs whose relative engagement order is correctly preserved. nDCG@10 evaluates retrieval quality by assessing whether the top-10 highest-engagement clips are correctly identified, with position-discounted weighting. Together, these metrics assess complementary aspects of how well each representation aligns with the full ordinal structure of the engagement signal, rather than merely distinguishing high from low extremes. This analysis also serves to motivate the selection of representations for the binary classification experiments that follow.

We report results for the primary set of representations in Table~\ref{tab:correlation}. Extended comparisons including additional VideoMAE sampling densities, weighted averaging variants, and segment-level breakdowns are provided in Supplementary Section~S2, all of which yield consistent conclusions. Across all four metrics, representations follow a clear ordering along the temporal richness continuum: video features achieve the highest correlation ($\rho = 0.626$, skip-mid $\rho = 0.785$ for VideoMAE-16), followed by TAIs ($\rho = 0.545$, skip-mid $\rho = 0.703$), text embeddings ($\rho = 0.517$), mid-point frames ($\rho = 0.496$), and audio ($\rho = 0.205$). The ordering of TAIs above mid-point frames, despite both using the same MAE ViT-L backbone, indicates that temporal compression itself contributes to alignment, beyond what any single frame can provide. All representations substantially exceed the random baseline ($\rho \approx -0.002$), confirming genuine alignment with the engagement signal.

Excluding clips near the median heat value (skip-mid analysis) improves correlations across all representations while preserving the same ordering, suggesting that the ambiguous middle range of the engagement distribution is consistently harder to predict across representational forms.

\begin{table*}[htbp]
    \centering
    \scriptsize
    \setlength{\tabcolsep}{3pt}
    \renewcommand{\arraystretch}{0.9}
    \caption{Ranking-based alignment metrics with replay heat values (mean across 61 videos). Skip-mid $\rho$ excludes clips within 30\% of the median heat value. All representations substantially exceed the random baseline.}
    \label{tab:correlation}
        \begin{tabular}{lccccc}
            \toprule
            \textbf{Representation} & \textbf{Spearman $\rho$} & \textbf{Skip-mid $\rho$} & \textbf{Kendall $\tau$} & \textbf{Pairwise Acc.} & \textbf{nDCG@10} \\
            \midrule
            Video (VideoMAE-16)            & 0.626 & 0.785 & 0.463 & 0.731 & 0.875 \\
            TAI (MAE ViT-L)                & 0.545 & 0.703 & 0.395 & 0.698 & 0.812 \\
            Text (EVA-CLIP)                & 0.517 & 0.637 & 0.373 & 0.686 & 0.728 \\
            Mid-point frame (MAE ViT-L)    & 0.496 & 0.671 & 0.352 & 0.675 & 0.756 \\
            Audio (VGGish)                 & 0.205 & 0.298 & 0.139 & 0.569 & 0.499 \\
            \midrule
            Random baseline                & $-$0.002 & 0.009 & $-$0.001 & 0.500 & 0.384 \\
            \bottomrule
        \end{tabular}
\end{table*}

\subsection{Binary Classification}\label{ssec:binary-classification}

Building on the correlation analysis, which establishes that video features and TAIs occupy the two ends of the performance range among visual representations, we focus the binary classification evaluation on these two modalities, with audio and text included as additional baselines. Classification setup and classifier selection are described in the Methods section. Full results across all classifiers and thresholds are reported in Supplementary Table~S3.

\begin{figure*}[htbp]
    \centering
    \includegraphics[width=1\textwidth]{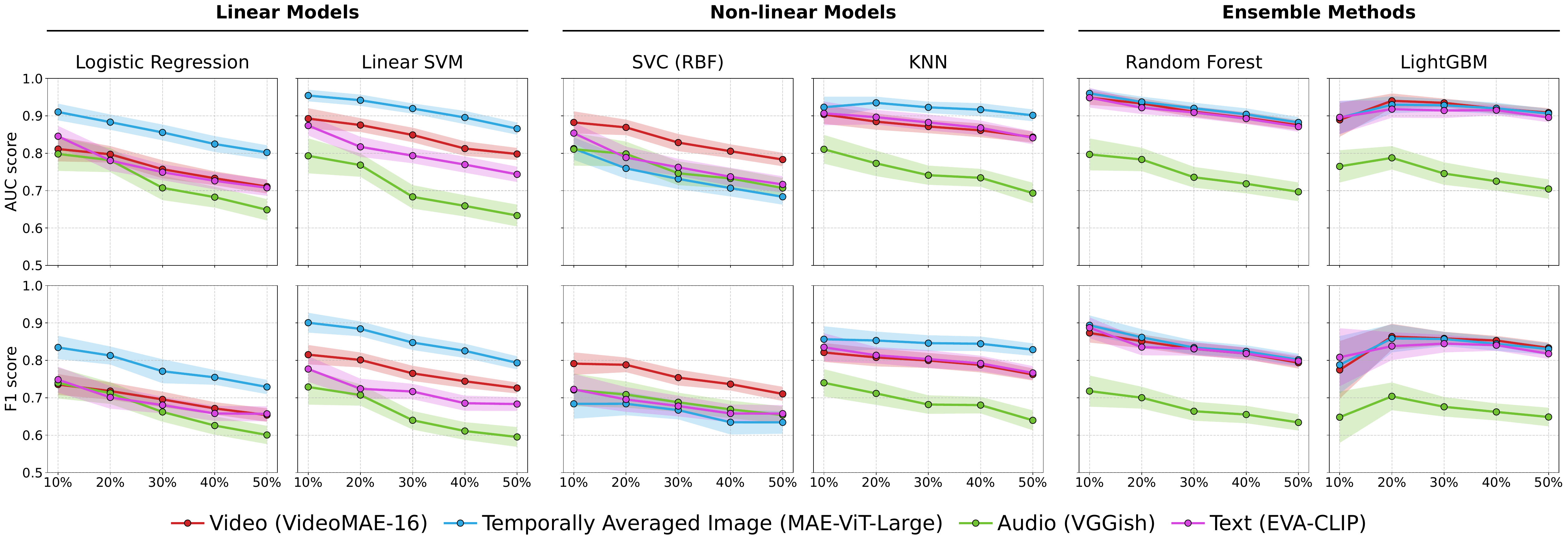}
    \caption{Model-wise AUC (top) and F1 (bottom) across quantile splits for different feature representations. \textbf{Solid lines represent mean values, with shaded areas indicating 95\% confidence intervals (CIs).} Despite preserving fine-grained temporal information, video-based representations do not consistently outperform temporally compressed images across classifiers and thresholds. This pattern suggests that increased representational richness alone does not guarantee closer alignment with human engagement signals.}
    \label{fig:binary}
\end{figure*}

Across linear and nonlinear classifiers, TAI features consistently outperform spatiotemporal video features (Fig.~\ref{fig:binary}). Under Logistic Regression at the 10\% quantile threshold, TAIs achieve an AUC of 0.910 and F1 of 0.835, compared to 0.811 and 0.735 for VideoMAE-16. Under Linear SVC, the margin is even larger: TAIs achieve an AUC of 0.954, compared with 0.893 for video. KNN follows the same pattern, with TAIs outperforming video across all five quantile thresholds.

Under ensemble methods, the performance gap between TAIs and video features largely disappears. Random Forest yields nearly identical results for both representations across all thresholds (e.g., AUC 0.960 vs. 0.948 at 10\%). LightGBM shows a similarly close pattern, with the two representations alternating marginal leads across quantiles and no consistent directional advantage emerging for either. Taken together, ensemble classifiers do not reveal a systematic superiority of video features. Rather, they confirm that the discriminative information captured by TAIs is broadly comparable to that of full spatiotemporal representations, even under models capable of exploiting high-dimensional feature interactions.

SVC with RBF kernel is the one consistent exception, where video features outperform TAIs across all quantile thresholds (e.g., AUC 0.882 vs. 0.812 at 10\%). We attribute this to the sensitivity of kernel-based methods to feature distributions. The motion-blurred appearance of TAIs deviates from the clean natural image statistics on which MAE ViT-L was pretrained, resulting in a feature space that is less amenable to RBF kernel separation. Notably, even under this unfavorable condition, TAIs retain competitive absolute performance.

Audio-based representations consistently yield the lowest scores across all classifiers and thresholds, a pattern reinforced by their near-random Spearman correlation ($\rho = 0.205$). Acoustic information contributes minimally to engagement prediction, consistent with the predominantly visual nature of egocentric urban experience. Sounds provide contextual background but do not carry the compositional cues that drive viewer attention.

Text embeddings perform substantially better than audio and, in the correlation analysis, outperform even single mid-point frames ($\rho = 0.517$ vs.\ $0.496$). Their predictive signal reflects the correlation between high-level semantic scene descriptions and engagement. One important caveat is that the text features here are not a direct sensory channel. They are generated by first extracting frame-wise keywords via a vision-language model and then re-encoding those descriptions as semantic embeddings. This two-stage pipeline is an analytical reconstruction of visual content rather than anything pedestrians directly experience while walking. Their reasonable performance is therefore best understood as evidence that semantic scene content carries engagement-relevant signal, not that language mediates urban experience. The human validation result, discussed below, reinforces this distinction.

The robustness of the video-versus-TAI comparison across three temporal sampling densities (16, 64, and 256 frames) confirms that these patterns reflect representational properties rather than frame sparsity (Supplementary Section~S2.1).

Taken together, these results indicate that TAIs capture the majority of the discriminative signal relevant to moment-level engagement prediction, and that additional temporal detail does not uniformly translate into improved classification performance.

\subsection{Human Validation}\label{ssec:human-validation}

To assess whether replay-based engagement signals correspond to explicit human judgment, we conducted an independent two-alternative forced-choice (2AFC) study on Amazon Mechanical Turk (AMT). Workers were presented with pairs of clips drawn from the same source video and asked to select the option that appeared more engaging or more representative of the street-walking experience. Each pair consisted of one high-heat and one low-heat clip with a minimum heat gap of 0.3, ensuring that engagement differences were sufficiently distinct to be perceptible. Three parallel tasks were deployed corresponding to the three representational media: temporally averaged images, 10-second video clips, and keyword-based text descriptions. The clip with the higher replay heat value served as the reference label, enabling direct comparison between explicit human preference and the implicit engagement signal. Full details of the annotation protocol are provided in the Methods section.

\begin{figure}[htbp]
    \centering
    \includegraphics[width=1\columnwidth]{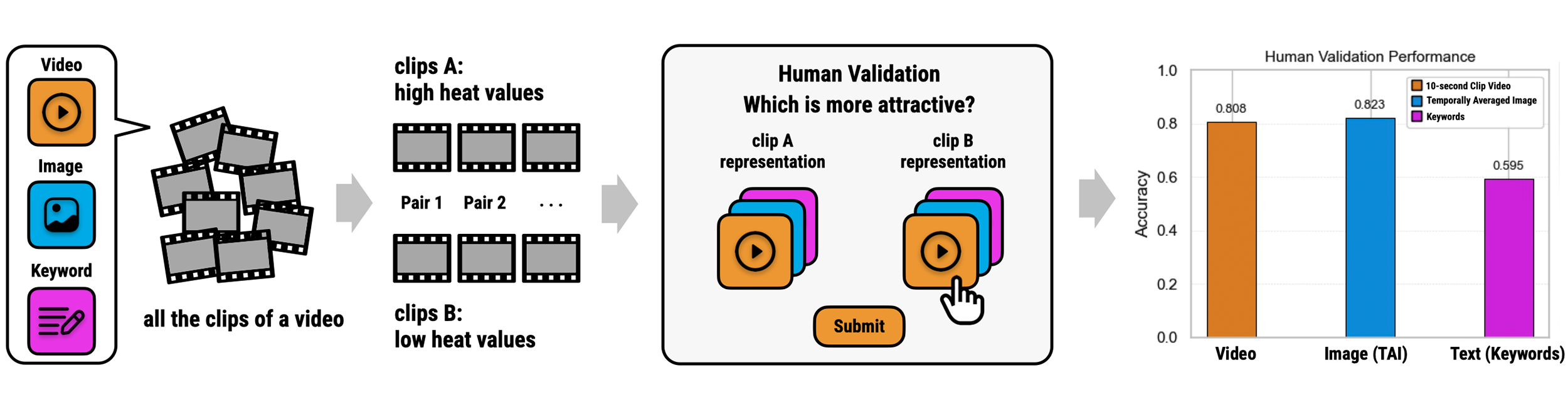}
    \caption{Overview of the human validation pipeline. From left to right: multimodal representations (video, TAI, and keywords); construction of within-video high-heat vs. low-heat pairs; two-alternative forced-choice (2AFC) evaluation on AMT; and aggregated accuracies across representational conditions.}
    \label{fig:human-validation-pipeline}
\end{figure}

Across representational conditions, human judgments showed systematic correspondence with the replay-based engagement signal (Fig.~\ref{fig:human-validation-pipeline}). Using TAIs alone, participants correctly identified the higher-heat moment with an overall accuracy of 86.38\%, comparable to the full video condition (85.71\%). This parity indicates that the additional temporal information present in video clips did not yield a discriminative advantage over the compressed visual summary. In contrast, text-based descriptions yielded substantially lower accuracy (65.12\%), a gap of roughly 20 percentage points relative to both visual conditions. When language descriptions serve as the sole input, participants lose access to the visual and temporal cues that most reliably distinguish engaging from non-engaging moments. Text features may carry engagement-relevant semantic signal, but they do not substitute for perceptual experience. They capture what a scene can be described as, not what it feels like to move through it.

To examine robustness under more ambiguous conditions, we further analyzed pairs with smaller heat differences (0.3--0.5), comprising approximately 15\% of total pairs ($N\approx 46$ per condition). In this more difficult regime, TAIs maintained high discriminative accuracy (88.6\%), matching or surpassing the video condition (85.7\%). Together, these results confirm that the alignment observed in binary classification reflects genuine human perceptual correspondence rather than an artifact of the replay-based signal, and that temporally compressed visual summaries retain sufficient information for humans to distinguish relative engagement across a wide range of everyday street scenes.

\subsection{Representational Divergence Analysis}\label{ssec:gap-analysis}

To better understand when temporally rich video representations offer advantages over compressed visual summaries, and when they do not, we conducted a gap analysis focusing on moments where the two modalities diverged in predictive performance. We base the analysis on predictions from the Linear SVC classifier, using VideoMAE-16 features for video and MAE ViT-L features for TAIs, both of which exhibited stable performance in the binary classification task. For each 10-second clip, we record whether the prediction based on video features and the prediction based on TAI features are correct with respect to the ground-truth engagement label. We define two contrastive groups:

\begin{itemize}
    \item Group A ($\mathcal{G}_A$): clips for which the TAI prediction is correct while the video prediction is incorrect;
    \item Group B ($\mathcal{G}_B$): clips for which the video prediction is correct while the TAI prediction is incorrect.
\end{itemize}

Clips for which both representations are correct, or both are incorrect, are excluded, isolating cases in which representational choice rather than overall task difficulty drives differences in predictive alignment (Supplementary Table~S5). Notably, $\mathcal{G}_A$ is substantially larger than $\mathcal{G}_B$ (463 vs.\ 182), consistent with the binary classification finding that TAIs more frequently match the engagement signal.

A semantic segmentation analysis further characterizes the two groups at the scene level (full results in Supplementary Section~S3.2). The most consistent distinction is temporal: $\mathcal{G}_B$ clips exhibit markedly higher frame-to-frame variance in building, sky, and person proportions, indicating greater scene dynamics. Among mean descriptors, the clearest difference appears in low-score clips, where $\mathcal{G}_A$ shows notably higher greenery coverage than $\mathcal{G}_B$ (0.10 vs.\ 0.06), while other mean quantities remain broadly similar across groups.

\begin{figure*}[t]
    \centering
    \begin{subfigure}[t]{0.49\columnwidth}
        \centering
        \includegraphics[width=\linewidth]{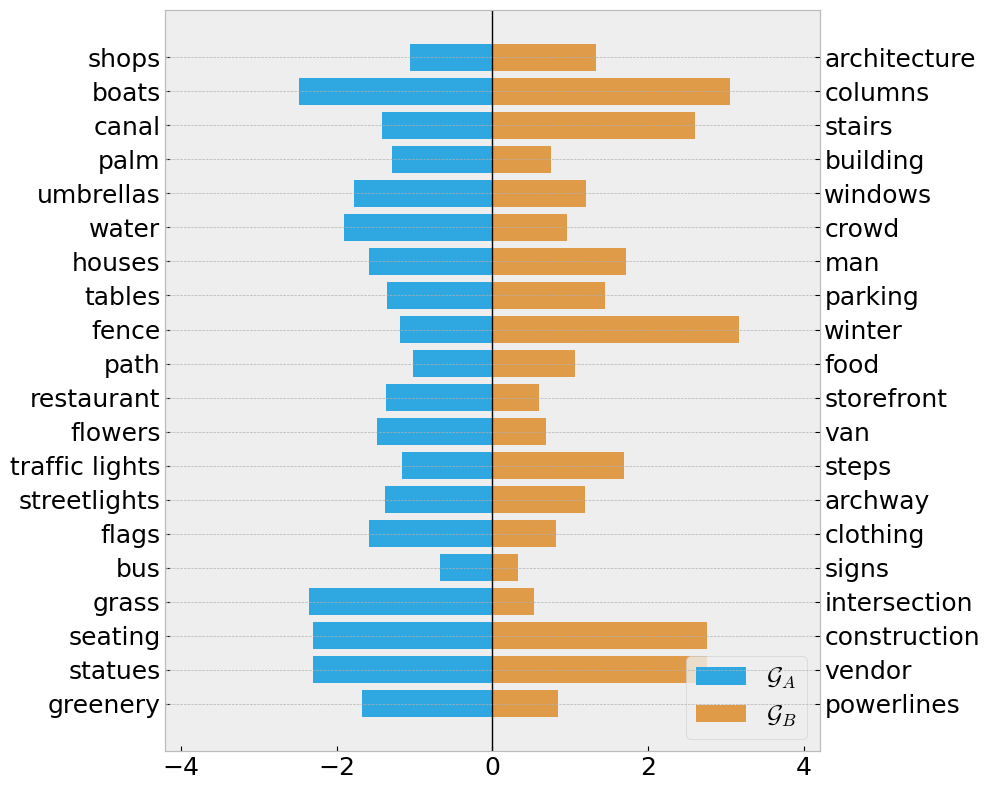}
        \caption{High-score clips}
        \label{fig:high-comp}
    \end{subfigure}
    \hfill
    \begin{subfigure}[t]{0.49\columnwidth}
        \centering
        \includegraphics[width=\linewidth]{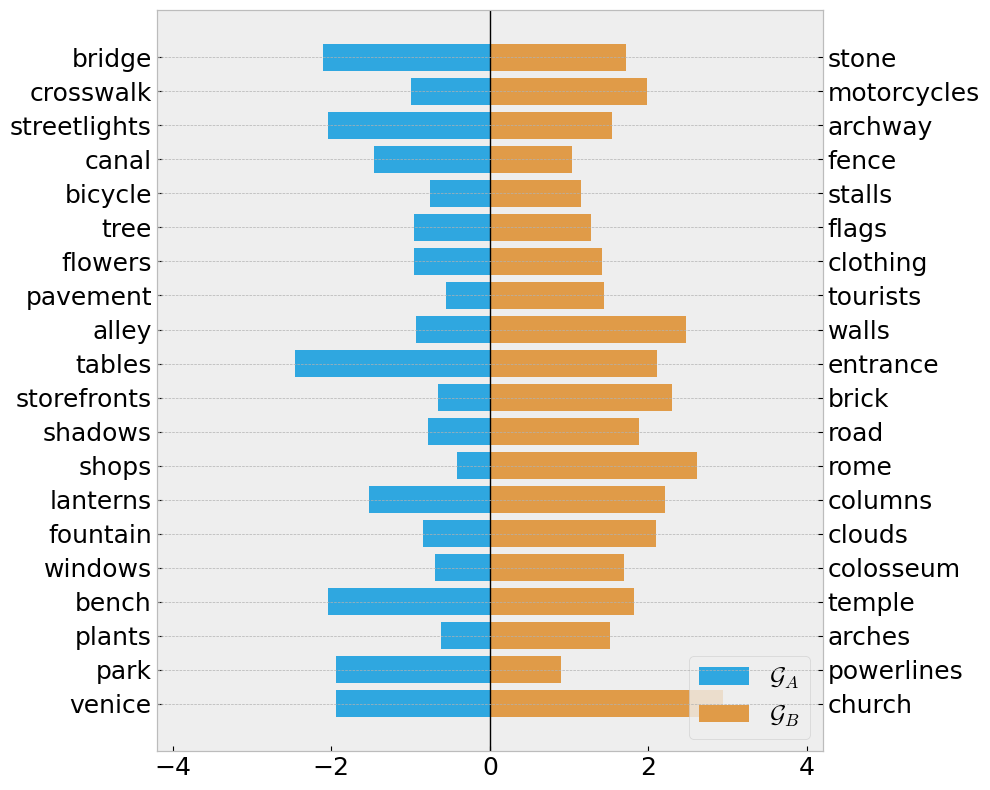}
        \caption{Low-score clips}
        \label{fig:low-comp}
    \end{subfigure}
    \caption{
        Comparative analysis of top-20 distinctive keywords for $\mathcal{G}_A$ (where TAI excels) and $\mathcal{G}_B$ (where video excels). Keywords are ranked by smoothed log-odds ratio \cite{monroe2008fightin}, with bar length indicating the degree of association with each group. Left (blue) bars represent features dominant in $\mathcal{G}_A$, while right (yellow) bars denote those characteristic of $\mathcal{G}_B$.
    }
    \label{fig:keyword-comp}
\end{figure*}

Because text features demonstrated meaningful alignment with engagement signals in the classification experiments, keyword distributions offer a productive diagnostic lens for characterizing what makes the two groups semantically distinct, even though language is not itself a direct perceptual channel. To examine high-level semantic content, we estimate group-specific keyword salience from frame-wise VLM descriptions using smoothed log-odds ratios with an informative Dirichlet prior~\cite{monroe2008fightin}, identifying terms disproportionately associated with each group while controlling for overall frequency effects (Fig.~\ref{fig:keyword-comp}).

The keyword distributions reveal a consistent semantic contrast between the two groups. $\mathcal{G}_A$ is characterized by natural and quotidian street elements: terms such as ``greenery,'' ``flowers,'' ``canal,'' and ``park'' in high-score clips, and ``crosswalk,'' ``pavement,'' ``alley,'' and ``bench'' in low-score clips, reflecting scenes where spatial composition and everyday street furniture dominate the visual field. $\mathcal{G}_B$, by contrast, is consistently associated with large-scale architectural and landmark vocabulary: ``architecture,'' ``columns,'' ``archway,'' and ``construction'' appear as distinctive across both score regions, alongside activity-related terms such as ``crowd,'' ``vendor,'' and ``storefront.'' Representative examples from each group are shown in Fig.~\ref{fig:clip-image-examples}. Additional examples for $\mathcal{G}_A$ and $\mathcal{G}_B$ are provided in Supplementary Figs.~S8 and~S9.

\begin{figure}[htbp]
    \centering
    \includegraphics[width=1\linewidth]{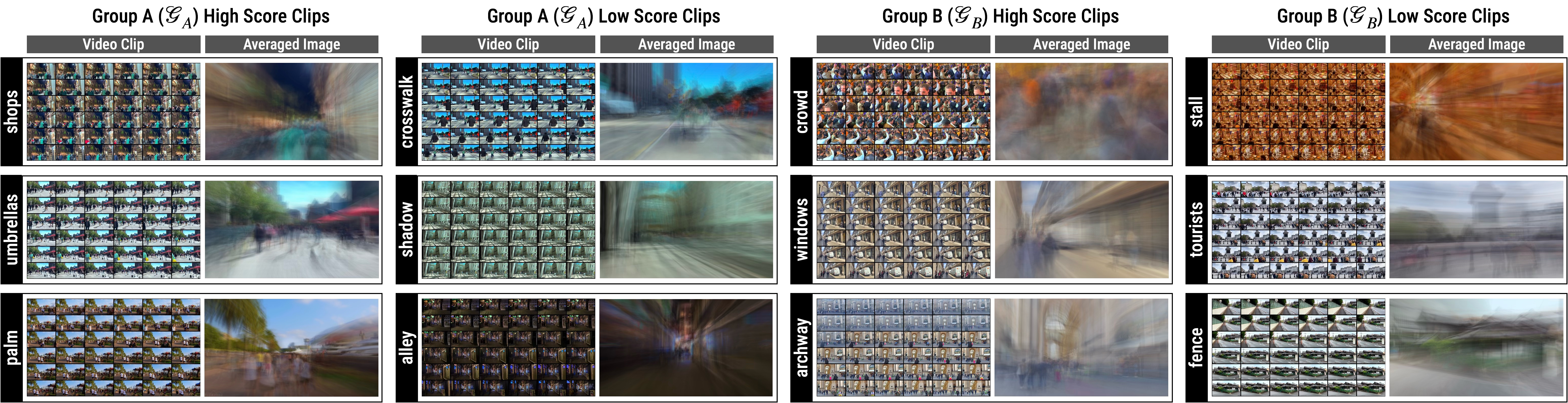}
    \caption{
        Illustrative examples from the four subgroups in the divergence analysis. Group A ($\mathcal{G}_A$) contains clips for which the TAI-based prediction is correct and the video-based prediction is incorrect, while Group B ($\mathcal{G}_B$) contains clips for which the video-based prediction is correct and the TAI-based prediction is incorrect. The four columns show, from left to right, $\mathcal{G}_A$ high-score clips, $\mathcal{G}_A$ low-score clips, $\mathcal{G}_B$ high-score clips, and $\mathcal{G}_B$ low-score clips. For each example, the left panel shows a film-strip visualization of the video clip and the right panel shows the corresponding temporally averaged image (TAI). $\mathcal{G}_A$ scenes are characterized by relatively stable spatial composition, whereas $\mathcal{G}_B$ scenes more often contain dynamic content that temporal averaging collapses into indistinct composites.
    }
    \label{fig:clip-image-examples}
\end{figure}

Crucially, the $\mathcal{G}_B$ keywords remain largely stable across high- and low-score clips, whereas $\mathcal{G}_A$ keywords shift considerably between the two score regions. This asymmetry carries an important interpretive implication: the scene content that distinguishes $\mathcal{G}_B$ (monumental architecture, structured urban corridors, and active street commerce) is not itself predictive of engagement level, yet video representations consistently succeed in these scenes regardless of score. This suggests that what video captures in $\mathcal{G}_B$ is not the static semantic content visible in keywords, but rather the temporal dynamics within the scene, including pedestrian flow, movement patterns, and activity rhythms, that TAIs collapse into indistinct composites. The keyword consistency of $\mathcal{G}_B$ thus reflects a representational dissociation: the scene \emph{type} is stable, but the engagement signal is carried by motion that only video can resolve.

\section{Discussion}\label{sec:discussion}

This study provides a controlled evaluation of how representation choice influences moment-level modeling of urban engagement. Across more than 50,000 street-level clips, the results challenge the implicit assumption that increasing representational richness yields more human-aligned measurement of urban experience.

\paragraph{From continuous alignment to binary prediction}
The reversal between correlation analysis (Section~\ref{ssec:correlation-analysis}) and binary classification (Section~\ref{ssec:binary-classification}) is the central empirical finding of this study. That video features outperform TAIs in continuous rank correlation is unsurprising, since richer temporal representations better track fine-grained ordinal variation across the full engagement distribution. What is noteworthy is that this advantage largely disappears under binary classification, the paradigm far more prevalent in urban perception research. Researchers in this field commonly assume that human perceptual responses are most reliable at the extremes, which is why pairwise comparison frameworks are used to collect labels and train scoring models that are then applied at scale. In this regime, the additional temporal detail that video preserves appears to contribute primarily to ordinal distinctions in the ambiguous middle range of the engagement distribution, rather than to the high-versus-low discrimination on which most practical pipelines depend. The AMT validation study (Section~\ref{ssec:human-validation}) confirms that this parity is not a modeling artifact: human participants showed comparable discriminative accuracy across TAI and video conditions, indicating that temporally compressed summaries retain sufficient information for the kind of evaluative judgment that underlies most large-scale urban perception research.

\paragraph{Exploring when representations diverge}
The exploratory gap analysis reported in Section~\ref{ssec:gap-analysis} is based on a single classifier and limited sample sizes, and is not intended as a definitive statistical test. Its purpose is to provide a diagnostic lens for understanding which environments favor one representation over the other. The observed pattern is nonetheless consistent: $\mathcal{G}_A$ clips tend to feature stable spatial compositions where TAIs effectively preserve the compositional cues relevant to engagement, while $\mathcal{G}_B$ clips are characterized by greater scene dynamics where temporal averaging collapses informative motion into indistinct composites. The keyword stability of $\mathcal{G}_B$ across high- and low-score clips is particularly telling: the scene type does not change, but the engagement signal does, suggesting that what video captures in these contexts is temporal dynamics rather than static semantic content. This pattern indicates that different representations are not simply better or worse in general, but capture different aspects of the urban environment. If the goal is to faithfully approximate human perceptual judgment, the choice between video and temporally compressed representations should be context-dependent rather than universal.

\paragraph{Selective compression in human perception}
The dissociation observed in the gap analysis points to a broader property of human perceptual judgment: the brain does not uniformly retain or discard temporal information, but appears to compress it selectively depending on the character of the scene. In spatially stable environments dominated by architectural structure, vegetation, and street geometry, observers seem to integrate visual input into a single coherent impression, effectively discarding moment-to-moment variation that adds little to the overall evaluation. This is the regime in which TAIs succeed, precisely because they mirror this compression. The mechanism is consistent with ensemble perception research \citep{whitney2018ensemble, manassi2022illusion}, which posits that the visual system extracts summary statistics over short temporal windows rather than encoding each moment independently. A related phenomenon in face perception reinforces this interpretation: averaged faces are consistently judged as more attractive and representative than any individual face, because compositing amplifies prototypical structure while suppressing idiosyncratic variation \citep{langlois1990attractive}.

In more dynamic environments, however, human perception becomes sensitive to fine-grained temporal cues that carry evaluative weight. Pedestrian density offers a clear example: moderate foot traffic may signal liveliness and social vitality, while excessive crowding produces discomfort, and the distinction between the two depends on temporal patterns of movement and spacing that a static or averaged representation cannot resolve. Video representations succeed in these cases because they preserve the temporal structure on which such judgments depend.

This selective compression highlights a limitation of current representational approaches. Existing pipelines operate in an all-or-nothing fashion, either discarding all temporal information or preserving it in full. Neither strategy mirrors the adaptive, context-dependent compression that characterizes human perception. Developing representations that selectively retain temporal detail where it matters while compressing where it does not remains an open challenge. At the scale of pedestrian urban experience, however, the composition-driven regime appears to predominate. Most everyday street scenes are defined by a relatively stable spatial structure, and the slow pace of walking further limits frame-to-frame variation. This baseline stability helps explain why TAIs perform comparably to video across the majority of clips, and why activity-driven scenes where video holds an advantage constitute a minority of the dataset.

\paragraph{Audio and text modalities}
Audio embeddings consistently show near-chance alignment with engagement signals, a result that is broadly expected: in environmental perception, vision dominates, and acoustic scene content contributes relatively little to how people evaluate streets, particularly when viewing video remotely.

Text-based representations occupy a more interesting position. That keyword-derived embeddings achieve reasonable classification performance despite being an entirely indirect reconstruction of visual content, generated by a VLM and re-encoded as semantic embeddings, speaks to the capacity of modern vision-language models to extract evaluatively meaningful scene descriptions. This suggests a form of cross-modal correspondence in which the semantic structure that language captures reflects aspects of visual experience relevant to human judgment. At the same time, the roughly 20-percentage-point accuracy gap between the text condition and both visual conditions in the AMT study confirms that language descriptions cannot substitute for direct perceptual access. Keywords capture what a scene can be described as, but not what it feels like to move through.

\paragraph{Implications for representation design}
These results highlight that representation choice acts as a filtering mechanism in computational urban analytics. Temporally compressed representations emphasize structural composition, while temporally rich representations amplify activity-related variation. As urban perception models are increasingly used to operationalize constructs such as liveliness, comfort, or attractiveness at scale, representational decisions may implicitly bias measurement toward particular dimensions of urban space without any explicit modeling choice to do so.

For practitioners, this implies that maximizing representational richness is not equivalent to improving measurement validity. Representation design should be aligned with the perceptual construct of interest. Tasks emphasizing social interaction, pedestrian dynamics, or short-lived events may benefit from temporal encoding, whereas evaluations centered on architectural character, spatial atmosphere, or vegetation quality may be adequately captured through temporally aggregated imagery at substantially lower computational cost. This distinction carries direct implications for sustainable urban planning: models used to evaluate walkability, green space quality, or neighborhood character should be matched to representations that preserve the perceptual cues most relevant to each construct---rather than defaulting to the richest available data, which may amplify transient activity signals at the expense of the stable spatial qualities that shape long-term livability.

\paragraph{Limitations}
Several limitations warrant consideration. Engagement is operationalized using replay heat values derived from a specific online platform, and although supported by human validation, this signal may still reflect platform-specific viewing behaviors. The dataset focuses on the ``city-walk'' genre, which carries consistent recording conventions that may not generalize to more heterogeneous street-level video sources or to perceptual constructs beyond general engagement, such as safety or comfort, which may exhibit different dependencies on temporal information. The gap analysis is based on a single classifier and limited sample sizes, and should be interpreted as exploratory rather than confirmatory. Future work should extend this framework to additional perceptual dimensions, data sources, and adaptive representational strategies that move beyond the current all-or-nothing paradigm.

Overall, this study reframes representation choice as a substantive methodological decision rather than a neutral preprocessing step. The results suggest that human perceptual judgment is itself selectively temporal: people compress or discard fine-grained variation in many everyday street contexts while relying on temporal dynamics in others. The choice of visual representation determines which perceptual regime a computational model captures, and empirical validation of this choice is essential when such models stand in for human judgment.

\section{Conclusion}\label{sec:conclusion}

We present a large-scale empirical evaluation of how representation choice shapes the modeling of street-level urban engagement. Across 61 city-walk videos and more than 50,000 clips, a Spearman correlation analysis confirms the expected ordering along the temporal richness continuum, with video features achieving the highest continuous alignment, followed by TAIs and single mid-point frames. However, under binary classification, the paradigm most widely used in urban perception research to train perceptual scoring models, TAIs consistently match or outperform video features, and an independent human validation study on Amazon Mechanical Turk confirms that this parity reflects genuine perceptual correspondence. An exploratory gap analysis suggests that this result arises from two complementary scene regimes: TAIs align more closely with engagement in composition-driven scenes dominated by stable spatial structure, while video representations are essential in activity-driven contexts where temporal dynamics carry the evaluative signal. Audio embeddings show near-chance alignment throughout, consistent with the dominance of vision in environmental perception, while text-based representations carry a moderate semantic signal but cannot substitute for direct visual access.

Together, these findings suggest that human perceptual judgment is itself selectively temporal. People compress fine-grained variation in many everyday street contexts while remaining sensitive to temporal dynamics in others. Current representational approaches do not mirror this adaptivity, operating instead in an all-or-nothing fashion. At the pedestrian scale, where stable spatial structure predominates and walking speed limits frame-to-frame variation, temporally compressed representations may not only be computationally cheaper but also more broadly aligned with human judgment. Representation design is a substantive methodological decision that determines which aspects of urban experience become legible to computational models, and empirical validation of this choice is essential when such models stand in for human judgment.

\section{Methods}\label{sec:methods}

\subsection{City-walk Video Dataset}\label{ssec:data-collection}

To evaluate representational effects under naturalistic pedestrian conditions, we curated the \textbf{\dataset}, a collection of 61 first-person city-walk videos sourced from publicly available YouTube content. This genre typically depicts continuous forward motion along urban streets and sidewalks with minimal editing, thereby approximating an egocentric street-level visual experience.

\paragraph{Initial collection}
We conducted video collection in two rounds during 2023 using keyword-based search queries related to city walking and urban exploration. The initial pool contained 117 videos spanning a wide range of cities, recording styles, and production qualities. To maintain a consistent pedestrian perspective, we excluded videos dominated by vehicular viewpoints or aerial footage (e.g., drone-based), reducing the pool to 82 videos.

\paragraph{Engagement-based filtering}
Because our analysis relies on replay-based heat values as an engagement signal, we further filtered videos based on the availability and variability of these signals. We retained only videos whose cumulative heat values exceeded a minimum threshold to ensure sufficient within-video variation, and excluded source videos that yielded fewer than 100 valid clips after cleaning. This step reduced the dataset to 61 videos, totaling approximately 109 hours of footage.

\begin{figure}[htbp]
\centering
\includegraphics[width=\linewidth]{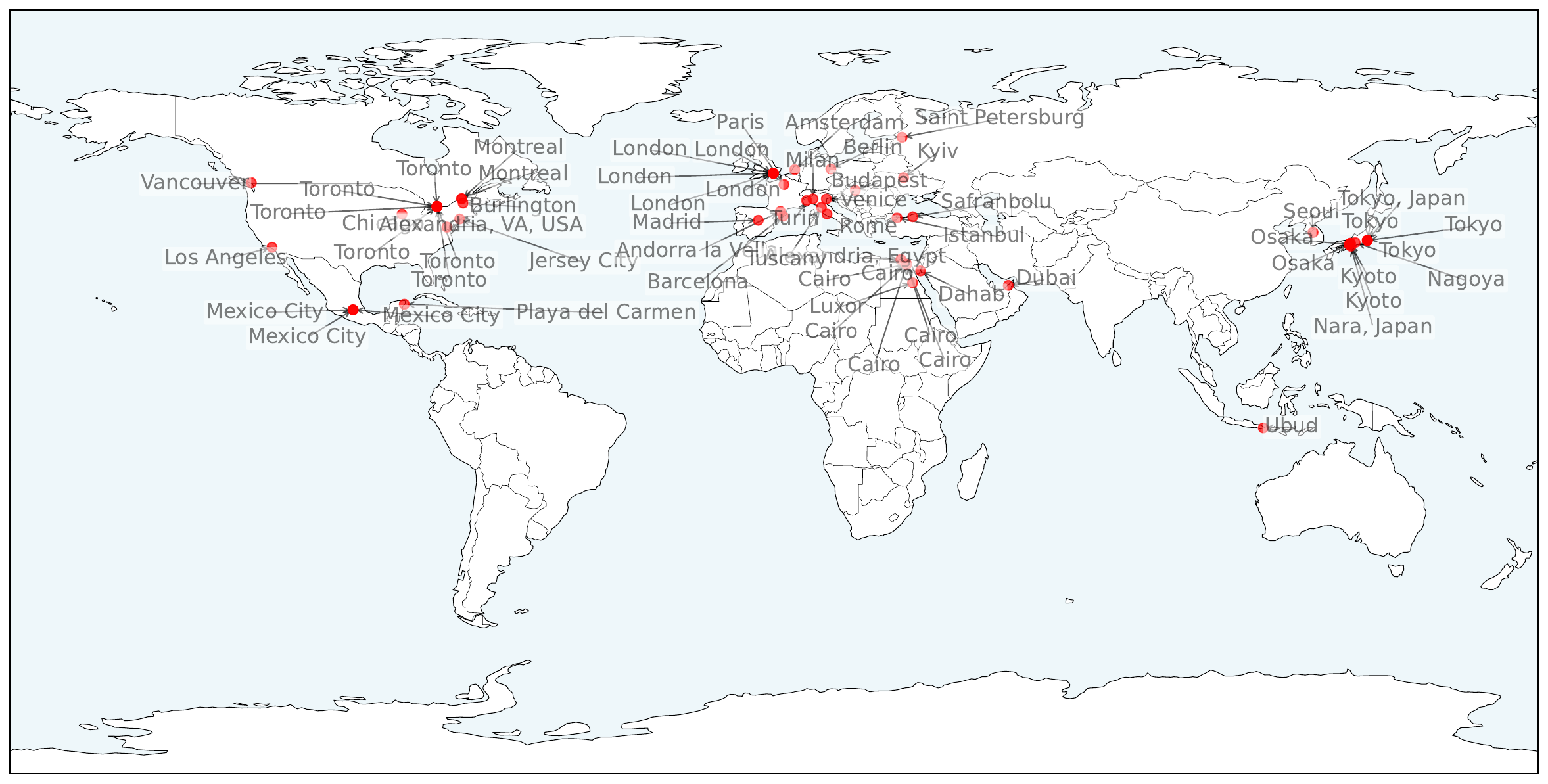}
\caption{Geographic distribution of the \dataset. Red markers indicate cities represented in the dataset, spanning North America, Europe, Asia, and the Middle East. The diversity of urban contexts supports the generalizability of findings across different street environments and cultural settings.}
\label{fig:geodist}
\end{figure}

The resulting \dataset spans a geographically diverse set of urban environments, as shown in Fig.~\ref{fig:geodist}.

\begin{figure}[htbp]
\centering
\includegraphics[width=1\columnwidth]{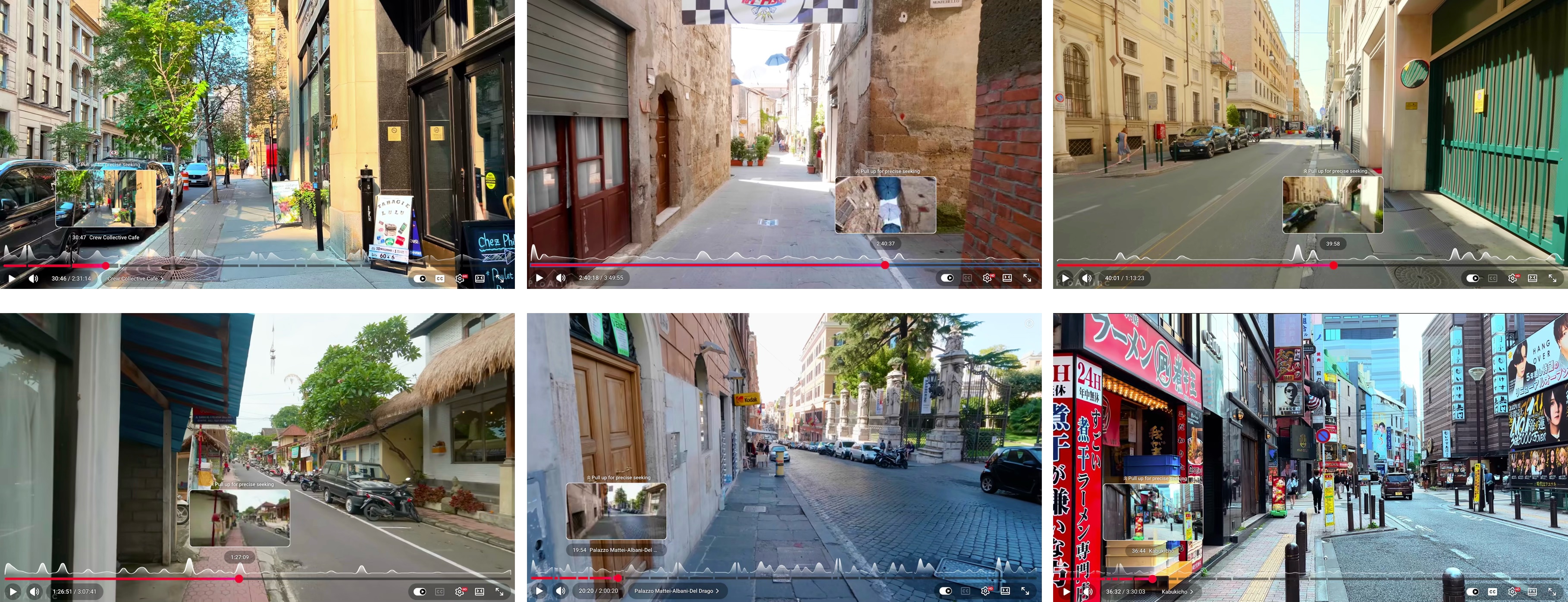}
\caption{Examples of pedestrian video footage with replay heat values. Each row shows a representative clip from the \dataset; the heat value curve (shown in the progress bar below each frame) reflects the relative frequency with which viewers replayed that segment, with peaks indicating moments of higher replay intensity. These values serve as the primary behavioral engagement signal throughout the study.}
\label{fig:replay-score}
\end{figure}

\subsection{Clip Segmentation}\label{ssec:data-clip}

All videos were segmented into overlapping 10-second clips with a 5-second stride. The 10-second window was chosen to approximate the temporal scale over which human observers form stable scene impressions \citep{whitney2018ensemble, manassi2022illusion}, while the 5-second stride ensures sufficient overlap to capture engagement transitions without introducing redundant clip pairs. After excluding the first and last 300 seconds of each recording to avoid platform-specific introductions and end screens, the corpus yielded approximately 71,000 candidate clips.

\subsection{Data Cleaning and Preprocessing}\label{ssec:data-cleaning}

To ensure that replay-based engagement signals reflect properties of street scenes rather than artifacts of video production, we applied two targeted filtering procedures prior to feature extraction.

\paragraph{Filtering editing transitions}
Some clips contain abrupt viewpoint changes, such as cuts between different streets, which may artificially inflate replay values by attracting viewer attention unrelated to the street scene itself. To identify such clips, we extracted one frame per second and computed frame-to-frame cosine similarity using ImageBind \citep{girdhar2023imagebind} features. Similarity scores were normalized to Z-scores within each video, and frames exceeding a threshold of 2 were flagged as potential transition points. A total of 24,652 frames were identified through this procedure. Any clip containing one or more flagged frames was excluded from subsequent analysis.

\paragraph{Removing graphic overlays and captions}
A second source of bias arises from embedded logos, watermarks, and overlaid captions, which can attract viewer attention independently of the street environment. We employed LLaVA 1.6 \citep{liu2023visual} to generate frame-level descriptions, and flagged frames containing keywords indicative of graphic overlays (e.g., ``watermark,'' ``overlay,'' ``subscribe''). A total of 612 frames were identified, and any clip containing such frames was removed. Representative examples of both filtering steps are shown in Supplementary Figs.~S1 and~S2.

After applying both filtering steps, the final dataset comprised \textbf{50,380} clean clips available for analysis.

\subsection{Replay Heat Signal Extraction}\label{ssec:data-heat}

YouTube provides a publicly visible replay heatmap for selected videos, visualizing aggregate viewer revisit behavior along the playback timeline. We use this signal as a scalable behavioral proxy for moment-level engagement. Heat values were extracted programmatically using \texttt{yt-dlp}, which retrieves video metadata including a heatmap field containing approximately 100 temporal anchor points per video. Each anchor point encodes a normalized replay-intensity value and its corresponding timestamp, thereby covering the full video duration.

\paragraph{Interpolation and per-second alignment}
To obtain a continuous engagement signal at per-second resolution, we fitted a smoothing spline to the anchor points using \texttt{UnivariateSpline} with a smoothing factor of $s=2$, which closely approximates the visual curve rendered on the YouTube progress bar. The fitted spline was then evaluated at each integer second of the video to produce a dense heat signal aligned with the per-second frame extraction used throughout the analysis (Supplementary Fig.~S3).

Clip-level engagement scores were computed by averaging the interpolated per-second heat values within each 10-second window, consistent with the overlapping segmentation described above. This procedure assigns a single scalar engagement score to each clip, which serves as the prediction target in all subsequent analyses.

Because replay behavior may reflect narrative structure or platform-specific dynamics in addition to perceptual salience, we treat replay heat as a weak supervisory signal that enables controlled comparison across representations, rather than as a definitive ground truth. Its relationship to explicit human preference is evaluated independently through the human annotation study described in the Human Annotation Protocol subsection.

\subsection{Feature Extraction Across Modalities}\label{ssec:feature-extraction}

For each clip, we extracted four parallel representations that vary in modality and temporal fidelity, spanning a continuum from full spatiotemporal video to single-frame snapshots.

\paragraph{Video features}
Extracted using a pretrained VideoMAE-v2 encoder \citep{wang2023videomae}, which encodes temporal dynamics across multiple sampled frames. We evaluated three temporal sampling densities (16, 64, and 256 frames) to assess robustness to frame resolution. Results across densities were consistent, and we report VideoMAE-16 as the primary video representation in the main text, with results for 64 and 256 frames provided in Supplementary Section~S2.1.

\paragraph{Temporally averaged image (TAI) features}
Frames were uniformly sampled at 1~Hz within each clip and averaged pixel-wise to produce a single composite image, hereafter referred to as a \textit{temporally averaged image} (TAI), compressing 10~seconds of visual input into a single representation. Visual features were then extracted using a pretrained MAE ViT-L backbone \citep{he2022masked}.

The TAI design is grounded in two converging lines of evidence. First, the same ensemble perception literature that motivated the 10-second clip window \citep{whitney2018ensemble, manassi2022illusion} also supports temporal averaging as a representational design choice. If the human visual system extracts summary statistics over short temporal windows rather than encoding each moment independently, pixel-wise frame averaging is a computational analog for that biological smoothing process, suppressing transient elements to reveal the stable scene schema that shapes remembered experience. Second, a closely related analogy comes from research on averaged faces: pixel-wise composites of many face images are consistently judged as more attractive and prototypical than any individual face, because averaging filters idiosyncratic variation and amplifies the stable structural features that the visual system uses to represent identity \citep{langlois1990attractive}. Applying the same logic to street-level pedestrian views, a TAI compresses an entire clip into a composite that foregrounds the dominant spatial composition, preserving the persistent configuration of buildings, greenery, and street furniture while attenuating transient elements. Using MAE ViT-L as the feature extractor ensures methodological consistency with the video encoder, as both are part of the same masked autoencoder family.

\paragraph{Mid-point frame features}
To establish a no-temporal baseline within the image modality, we also extract features from the single frame at the temporal midpoint of each clip using the same MAE ViT-L backbone. Unlike the TAI, which integrates information across the full clip duration, the mid-point frame carries no temporal information whatsoever. Comparing the two representations allows us to isolate the contribution of temporal compression itself.

\paragraph{Audio embeddings}
Extracted using the pretrained VGGish model, capturing acoustic properties of the street environment within each clip.

\paragraph{Text embeddings}
Frame-wise keywords were generated using Intern-VL3 \citep{wang2025internvl3} and encoded using an EVA-CLIP text encoder \citep{sun2023eva}. Keywords were aggregated at the clip level to produce a semantic representation of each clip. The following structured prompt was used to extract keywords from each video frame:

\begin{tcolorbox}
\footnotesize
\ttfamily
You are given a single video frame, which comes from a ``city-walk''-style video (a handheld or moving camera that simulates how a person experiences the street). Respond ONLY in strict JSON format, with no explanations or extra text:

\{
    ``walk\_view'': true $\mid$ false,  \% filmed from a walking perspective\\
    ``city\_street'': true $\mid$ false,  \% urban street or plaza\\
    ``keywords'': [``k1'', ``k2'', ``k3'', ``k4'', ``k5'']  \% exactly 5 short nouns
\}

\end{tcolorbox}

All encoders were used in their pretrained form without task-specific fine-tuning, ensuring that observed performance differences reflect representational properties rather than task-specific optimization.

\subsection{Classification Setup}\label{ssec:classification-setup}

To evaluate representational alignment with engagement signals, we formulated a within-video binary classification task. For each source video, clips were ranked by normalized replay heat values and assigned binary labels based on quantile thresholds, with clips in the upper quantile treated as high-engagement and those in the lower quantile as low-engagement. This within-video design controls for broad contextual differences across cities and channels, focusing the comparison on moment-level variation within the same visual sequence. Multiple quantile thresholds (10\%--50\%) were evaluated to assess robustness across engagement definitions.

We evaluated six classifiers spanning three paradigms: linear models including Logistic Regression and Linear Support Vector Classification (Linear SVC); nonlinear models comprising K-Nearest Neighbors (KNN) and SVC with an RBF kernel; and ensemble methods represented by Random Forest and LightGBM. These classifiers were chosen to span a range of inductive biases, from linear decision boundaries to kernel-based and tree-based nonlinear methods, enabling assessment of whether any representational advantage is classifier-specific. Identical train and test splits and evaluation protocols were applied across all representations, and no representation-specific hyperparameter tuning was performed.

\subsection{Human Annotation Protocol}\label{ssec:amt-protocol}

We conducted an independent human annotation study between December 2025 and January 2026 to obtain explicit judgments aligned with the engagement construct used in this paper. Three parallel tasks were deployed on Amazon Mechanical Turk (AMT), corresponding to the three representational media studied: keyword-based descriptions (text), temporally averaged images, and 10-second video clips. For each task, we collected 300 Human Intelligence Tasks (HITs), yielding three comparably sized sets of pairwise decisions.

\paragraph{Worker eligibility}
To minimize comprehension issues while maintaining a broad annotator pool, we did not require ``Master's'' status. Eligibility criteria included an all-time approval rate of at least 97\%, at least 10{,}000 previously approved HITs, and residency in predominantly English-speaking regions (United States, Australia, Canada, New Zealand, and the United Kingdom). These filters were chosen to balance annotation quality with sufficient throughput across the three tasks.

\paragraph{Task design}
Each HIT presented a pair of moments drawn from the same source video and asked workers to select the option that appeared more engaging or more representative of the street-walking experience. While the presentation format varied by representational medium, all tasks shared the same underlying decision rule to ensure comparability across conditions.

\emph{1) Content description (keywords):} Workers read two short keyword sets, each describing a different 10-second city-walk clip, and selected the set that suggested the more engaging scene. Typical completion time was 10--30\,s. Responses completed under 30\,s or over 5\,min may be rejected.

\emph{2) Image preference (TAIs):} Workers viewed two options, each rendered as a temporally averaged image alongside a representative frame, and selected the image suggesting the more engaging scene. Typical completion time was approximately 30\,s. Responses completed under 10\,s or over 5\,min may be rejected.

\emph{3) Video preference (10-second clips):} Workers watched two 10-second clips and selected the clip they found more engaging or more likely to rewatch. Instructions required workers to view both clips in full, with replay allowed as needed. Typical task time was approximately 1\,min. Submissions where videos were not fully watched, or that exceeded 5\,min, may be rejected.

For evaluation, the reference label for each pair was defined as the clip with the higher replay heat value, enabling direct comparison between explicit human judgments and the aggregate engagement signal used elsewhere in the study.

\paragraph{Quality control}
Two sources of noise were addressed during post-collection screening. First, responses completed in less than the minimum reasonable viewing or reading time were excluded: less than 30\,s for video and keyword tasks, less than 10\,s for image tasks. Second, to preserve diversity of viewpoints and mitigate individual bias, contributions beyond 50 tasks from the same worker were blocked or excluded. Example task interfaces for all three conditions are shown in Supplementary Fig.~S10.

Large language models (LLMs), specifically LLaVA~1.6~\cite{liu2023visual} and Intern-VL3~\cite{wang2025internvl3}, were used as tools for frame-level visual description and keyword extraction in the feature extraction pipeline described in Section~\ref{ssec:feature-extraction}. These models were used solely as data processing tools and did not contribute to the writing of this manuscript.

\section*{Ethics Statement}

The only human-participant component of this research was a validation survey conducted via Amazon Mechanical Turk. The study was reviewed and approved by the Committee on the Use of Humans as Experimental Subjects (COUHES) at the Massachusetts Institute of Technology under Protocol ID E-7326 and was determined to be exempt from full board review under Category 3 (Benign Behavioral Interventions) in accordance with 45 CFR 46.104(d)(3). All procedures were performed in accordance with relevant guidelines and regulations and with the Declaration of Helsinki. Informed consent was obtained from all participants prior to participation.

\section*{Data Availability}
The YouTube videos analyzed in this study are publicly available on the YouTube platform. The derived dataset of replay heat values, extracted clip-level features, and Amazon Mechanical Turk annotation results that support the findings of this study are not publicly deposited but are available from the corresponding author upon reasonable request.

\section*{Code Availability}
The custom code used for feature extraction, classification, and analysis in this study is available from the corresponding author upon reasonable request.

\section*{Funding}
This work used Indiana Jetstream2 GPU resources through ACCESS allocation CIS250360, which is supported by National Science Foundation grants \#2138259, \#2138286, \#2138307, \#2137603, and \#2138296.

\section*{Acknowledgements}
We thank Fan Zhang and Chuanlin Lan for their contributions during the early exploratory phase of the project.

\section*{Author Contributions}
L.L. conceptualized the study, developed the methodology, curated the data, conducted the formal analysis, implemented the software, and prepared all visualizations. L.L. and F.H.T. wrote the original draft; F.H.T. led the design and execution of the Amazon Mechanical Turk validation study and wrote the human validation section. F.D. contributed to conceptualization and provided supervision. All authors reviewed and edited the manuscript.

\section*{Competing Interests}
All authors declare no financial or non-financial competing interests.

\section*{Additional Information}
\textbf{Supplementary information} is available for this paper.

\bibliographystyle{naturemag}
\bibliography{egbib}

\clearpage
\renewcommand{\thesection}{S\arabic{section}}
\renewcommand{\theHsection}{supp.S\arabic{section}}
\setcounter{section}{0}
\renewcommand{\theHsubsection}{supp.S\arabic{section}.\arabic{subsection}}
\renewcommand{\thefigure}{S\arabic{figure}}
\renewcommand{\theHfigure}{supp.S\arabic{figure}}
\setcounter{figure}{0}
\renewcommand{\thetable}{S\arabic{table}}
\renewcommand{\theHtable}{supp.S\arabic{table}}
\setcounter{table}{0}

\noindent {\Large \textbf{Supplementary Information}} \\
\noindent {\large \myTitle}

\vspace{1em}

\section{Details for Data and Feature Extraction}\label{sec:app-data-and-feature}

\subsection{Data Collection}\label{ssec:app-data-collection}

Full details of the data collection procedure, including initial pool construction, engagement-based filtering, and geographic coverage, are provided in the Methods section of the main text.

\begin{figure}[htbp]
    \centering
    \includegraphics[width=\linewidth]{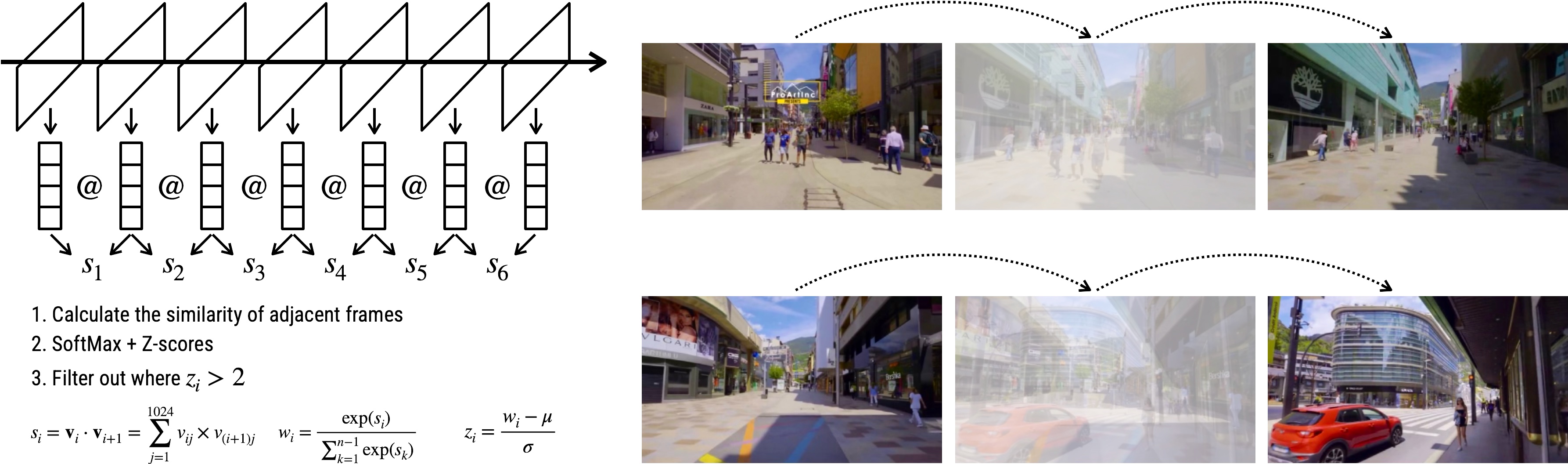}
    \caption{
        Detection of abrupt viewpoint changes via frame-to-frame visual dissimilarity. Each panel shows a flagged transition point: the left frame represents the last stable view before the cut, and the right frame shows the first frame after the transition. Cosine similarity scores (normalized within-video) are shown below each pair; flagged frames fall above the 2 threshold indicated by the dashed line.
    }
    \label{fig:transition-detection}
\end{figure}

\begin{figure}[htbp]
    \centering
    \includegraphics[width=\linewidth]{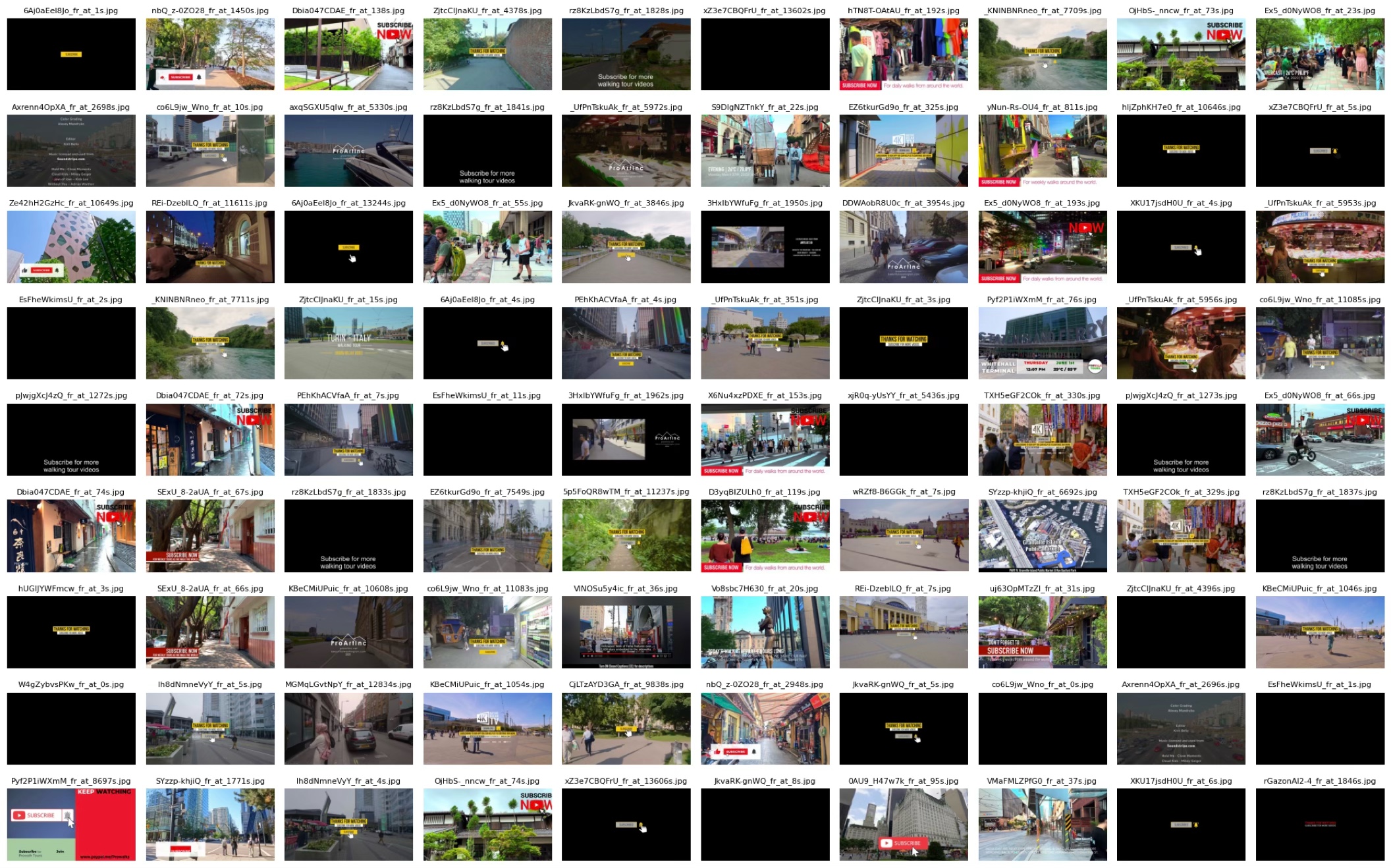}
    \caption{
        Examples of frames containing graphic overlays removed during preprocessing. Each panel shows a flagged frame alongside the detected keyword that triggered removal. Such elements attract viewer attention independently of the street environment and would distort replay-based engagement signals if retained.
    }
    \label{fig:overlay-detection}
\end{figure}

\subsection{Data Cleaning and Preprocessing}\label{ssec:app-preprocessing}

Full details of the data cleaning procedures, including editing transition detection and graphic overlay filtering, are provided in the Methods section of the main text. Representative examples of flagged frames are shown in Figs.~\ref{fig:transition-detection} and~\ref{fig:overlay-detection}.

\subsection{Extraction of Heat Values}\label{ssec:app-heat-value}

\begin{figure}[htbp]
    \centering
    \includegraphics[width=\linewidth]{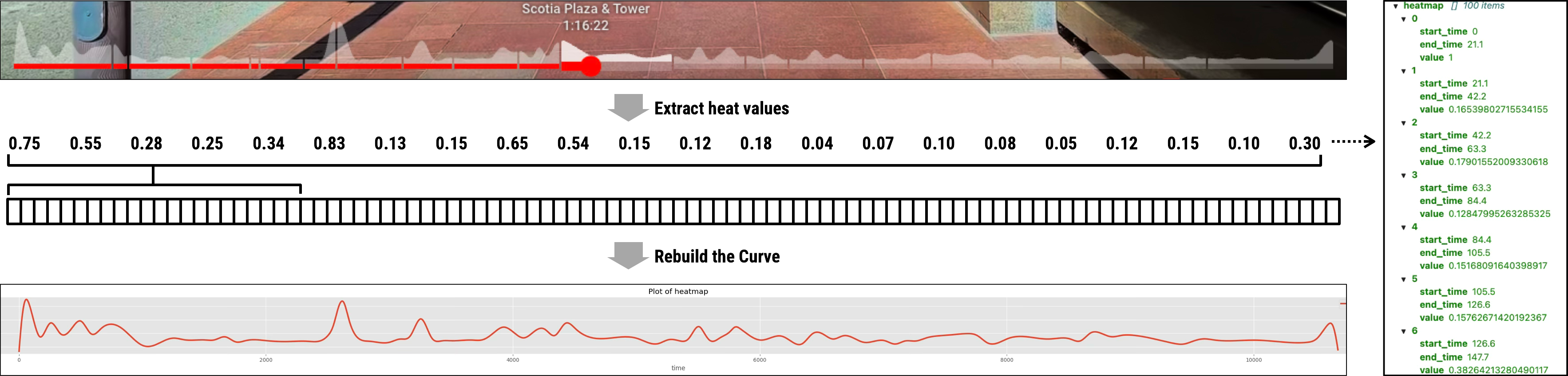}
    \caption{Heat value extraction and reconstruction pipeline. Top left: the replay heatmap as rendered on the YouTube progress bar, where the wave-like curve reflects aggregate viewer revisit frequency. Top right: the raw JSON output from \texttt{yt-dlp}, providing approximately 100 temporal anchor points each with a \texttt{start\_time}, \texttt{end\_time}, and normalized \texttt{value}. Bottom: the reconstructed continuous heat signal produced by fitting a smoothing spline ($s=2$) to the anchor points, closely approximating the original YouTube curve.}
    \label{fig:heat-extraction}
\end{figure}

Full details of the heat value extraction and spline interpolation procedure are provided in the Methods section of the main text. The extraction pipeline is illustrated in Fig.~\ref{fig:heat-extraction}.

\subsection{Vision-Language Model Prompt}\label{ssec:app-prompt}

\begin{figure}[htbp]
    \centering
    \includegraphics[width=\linewidth]{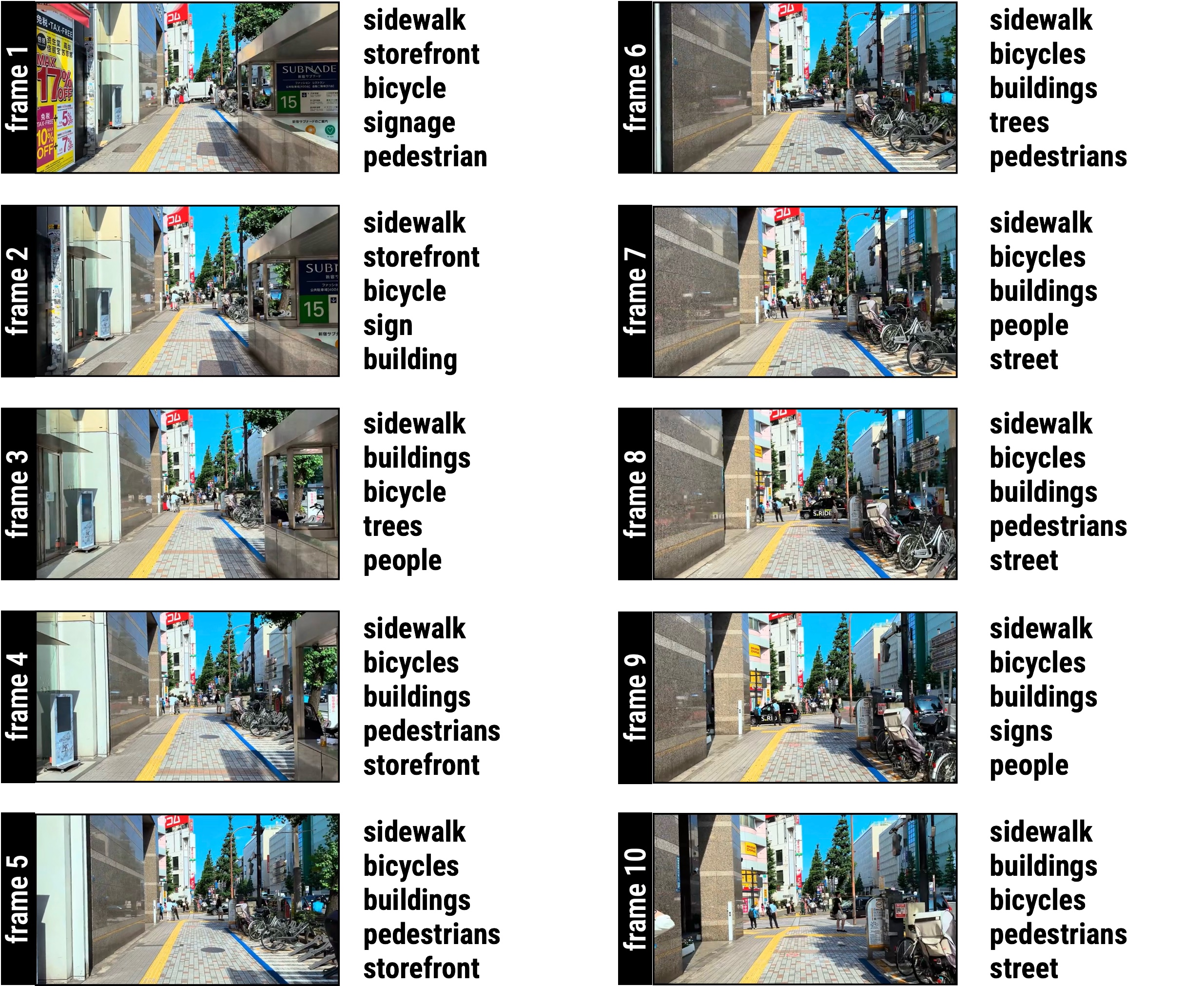}
    \caption{Example of frame-wise keyword extraction using the VLM prompt. For each video clip, we sample 10 frames (1\,Hz) and query the VLM with a structured JSON-based prompt to obtain five short nouns describing the visual content of each frame. The extracted keywords characterize the evolving scene semantics over time and are subsequently aggregated into clip-level keyword distributions for log-odds analysis.}
    \label{fig:prompt-frame}
\end{figure}

The VLM prompt used for frame-wise keyword extraction is provided in full in the Methods section of the main text. Fig.~\ref{fig:prompt-frame} illustrates how a clip can be understood in terms of frame-wise keywords.

\section{Details for Correlation and Classification}\label{sec:app-correlation-and-classification}

\subsection{VideoMAE Temporal Sampling Density}\label{ssec:app-videomae-frames}

Table~\ref{tab:spearman-videomae} reports Spearman $\rho$ and skip-mid $\rho$ for VideoMAE-16, 64, and 256 frames. Increasing frame density yields modest gains (from 0.626 to 0.684), but the ordering relative to TAIs is preserved across all densities: video features consistently outperform TAIs in continuous-valued alignment, while the margin does not grow substantially with additional frames.

\begin{table}[htbp]
    \centering
    \small
    \caption{Spearman correlation for VideoMAE under three temporal sampling densities.}
    \label{tab:spearman-videomae}
        \begin{tabular}{lcc}
            \toprule
            \textbf{Representation} & \textbf{Spearman $\rho$} & \textbf{Skip-mid $\rho$} \\
            \midrule
            Video (VideoMAE-16)  & 0.626 & 0.785 \\
            Video (VideoMAE-64)  & 0.675 & 0.796 \\
            Video (VideoMAE-256) & 0.684 & 0.805 \\
            \bottomrule
        \end{tabular}
\end{table}

\subsection{Weighted Averaging Variants}

Table~\ref{tab:spearman-weighted} reports results for two weighted variants of the TAI representation: a stability-weighted version that upweights frames with lower inter-frame variation, and a dynamic-weighted version that emphasizes frames with higher variation. Both variants yield results close to the plain TAI, with differences of less than 0.02 in Spearman's $\rho$, indicating that the choice of temporal weighting scheme does not materially affect alignment with engagement signals.

\begin{table}[htbp]
    \centering
    \small
    \caption{Spearman correlation for TAI weighting variants.}
    \label{tab:spearman-weighted}
        \begin{tabular}{lcc}
            \toprule
            \textbf{Representation} & \textbf{Spearman $\rho$} & \textbf{Skip-mid $\rho$} \\
            \midrule
            TAI (MAE ViT-L)              & 0.545 & 0.703 \\
            TAI, stability-weighted (MAE ViT-L)  & 0.563 & 0.724 \\
            TAI, dynamic-weighted (MAE ViT-L)    & 0.538 & 0.710 \\
            \bottomrule
        \end{tabular}
\end{table}

\subsection{Binary Classification: Full Results}\label{ssec:app-binary-full}

Table~\ref{tab:binary-detail} reports F1 and AUC for all six classifiers across five quantile thresholds and four feature types.

\begin{table}[htbp]
    \centering
    \scriptsize
    \setlength{\tabcolsep}{3pt}
    \renewcommand{\arraystretch}{0.9}
    \caption{Binary classification performance (F1 and AUC) across classifiers, quantile thresholds, and feature types. Bold indicates the best-performing representation for each classifier--quantile combination.}
    \label{tab:binary-detail}
    \resizebox{\columnwidth}{!}{
    \begin{tabular}{llcccccccc}
        \toprule
        & & \multicolumn{2}{c}{\textbf{Video}}
          & \multicolumn{2}{c}{\textbf{TAI}}
          & \multicolumn{2}{c}{\textbf{Audio}}
          & \multicolumn{2}{c}{\textbf{Text}} \\
        \cmidrule(lr){3-4} \cmidrule(lr){5-6} \cmidrule(lr){7-8} \cmidrule(lr){9-10}
        \textbf{Classifier} & \textbf{Quantile}
        & \textbf{AUC} & \textbf{F1}
        & \textbf{AUC} & \textbf{F1}
        & \textbf{AUC} & \textbf{F1}
        & \textbf{AUC} & \textbf{F1} \\
        \midrule
        \multirow{5}*{Logistic Regression} & 10\% & 0.811 & 0.735 & \textbf{0.910} & \textbf{0.835} & 0.798 & 0.739 & 0.845 & 0.748 \\
        & 20\% & 0.797 & 0.718 & \textbf{0.883} & \textbf{0.813} & 0.782 & 0.712 & 0.780 & 0.701 \\
        & 30\% & 0.757 & 0.696 & \textbf{0.856} & \textbf{0.771} & 0.707 & 0.662 & 0.749 & 0.680 \\
        & 40\% & 0.733 & 0.671 & \textbf{0.825} & \textbf{0.755} & 0.683 & 0.626 & 0.726 & 0.658 \\
        & 50\% & 0.711 & 0.654 & \textbf{0.802} & \textbf{0.729} & 0.649 & 0.600 & 0.708 & 0.657 \\
        \midrule
        \multirow{5}*{Linear SVC} & 10\% & 0.893 & 0.815 & \textbf{0.954} & \textbf{0.901} & 0.793 & 0.729 & 0.874 & 0.777 \\
        & 20\% & 0.875 & 0.801 & \textbf{0.942} & \textbf{0.884} & 0.768 & 0.707 & 0.817 & 0.724 \\
        & 30\% & 0.849 & 0.765 & \textbf{0.920} & \textbf{0.848} & 0.683 & 0.640 & 0.793 & 0.717 \\
        & 40\% & 0.812 & 0.744 & \textbf{0.895} & \textbf{0.825} & 0.659 & 0.611 & 0.770 & 0.685 \\
        & 50\% & 0.798 & 0.726 & \textbf{0.866} & \textbf{0.793} & 0.633 & 0.595 & 0.744 & 0.683 \\
        \midrule
        \multirow{5}*{SVC (RBF)} & 10\% & \textbf{0.882} & \textbf{0.791} & 0.812 & 0.684 & 0.811 & 0.721 & 0.854 & 0.722 \\
        & 20\% & \textbf{0.869} & \textbf{0.788} & 0.759 & 0.684 & 0.798 & 0.709 & 0.788 & 0.695 \\
        & 30\% & \textbf{0.829} & \textbf{0.754} & 0.731 & 0.667 & 0.746 & 0.688 & 0.762 & 0.678 \\
        & 40\% & \textbf{0.805} & \textbf{0.736} & 0.707 & 0.635 & 0.733 & 0.669 & 0.737 & 0.658 \\
        & 50\% & \textbf{0.783} & \textbf{0.710} & 0.684 & 0.634 & 0.707 & 0.654 & 0.717 & 0.658 \\
        \midrule
        \multirow{5}*{KNN} & 10\% & 0.904 & 0.821 & \textbf{0.923} & \textbf{0.856} & 0.811 & 0.740 & 0.907 & 0.835 \\
        & 20\% & 0.884 & 0.808 & \textbf{0.935} & \textbf{0.853} & 0.773 & 0.712 & 0.896 & 0.814 \\
        & 30\% & 0.871 & 0.800 & \textbf{0.923} & \textbf{0.846} & 0.741 & 0.682 & 0.882 & 0.803 \\
        & 40\% & 0.861 & 0.788 & \textbf{0.917} & \textbf{0.844} & 0.734 & 0.680 & 0.868 & 0.792 \\
        & 50\% & 0.843 & 0.762 & \textbf{0.901} & \textbf{0.829} & 0.693 & 0.640 & 0.841 & 0.766 \\
        \midrule
        \multirow{5}*{Random Forest} & 10\% & 0.948 & 0.873 & \textbf{0.960} & \textbf{0.894} & 0.797 & 0.718 & 0.948 & 0.887 \\
        & 20\% & 0.933 & 0.851 & \textbf{0.937} & \textbf{0.862} & 0.783 & 0.700 & 0.922 & 0.835 \\
        & 30\% & 0.911 & 0.831 & \textbf{0.920} & \textbf{0.834} & 0.736 & 0.664 & 0.909 & 0.831 \\
        & 40\% & 0.895 & 0.818 & \textbf{0.904} & \textbf{0.824} & 0.718 & 0.655 & 0.893 & 0.818 \\
        & 50\% & 0.875 & 0.794 & \textbf{0.882} & \textbf{0.802} & 0.697 & 0.634 & 0.871 & 0.798 \\
        \midrule
        \multirow{5}*{LightGBM} & 10\% & 0.889 & 0.774 & 0.895 & 0.788 & 0.765 & 0.648 & \textbf{0.896} & \textbf{0.808} \\
        & 20\% & \textbf{0.940} & \textbf{0.864} & 0.930 & 0.859 & 0.788 & 0.704 & 0.918 & 0.838 \\
        & 30\% & \textbf{0.935} & \textbf{0.859} & 0.928 & 0.857 & 0.746 & 0.676 & 0.914 & 0.845 \\
        & 40\% & 0.919 & \textbf{0.853} & \textbf{0.920} & 0.844 & 0.725 & 0.662 & 0.915 & 0.841 \\
        & 50\% & \textbf{0.909} & \textbf{0.833} & 0.906 & 0.830 & 0.704 & 0.649 & 0.896 & 0.818 \\
        \bottomrule
    \end{tabular}
    }
\end{table}

\subsection{VideoMAE Temporal Sampling Sensitivity}\label{ssec:app-videomae-sensitivity}

\begin{figure}[htbp]
    \centering
    \includegraphics[width=\linewidth]{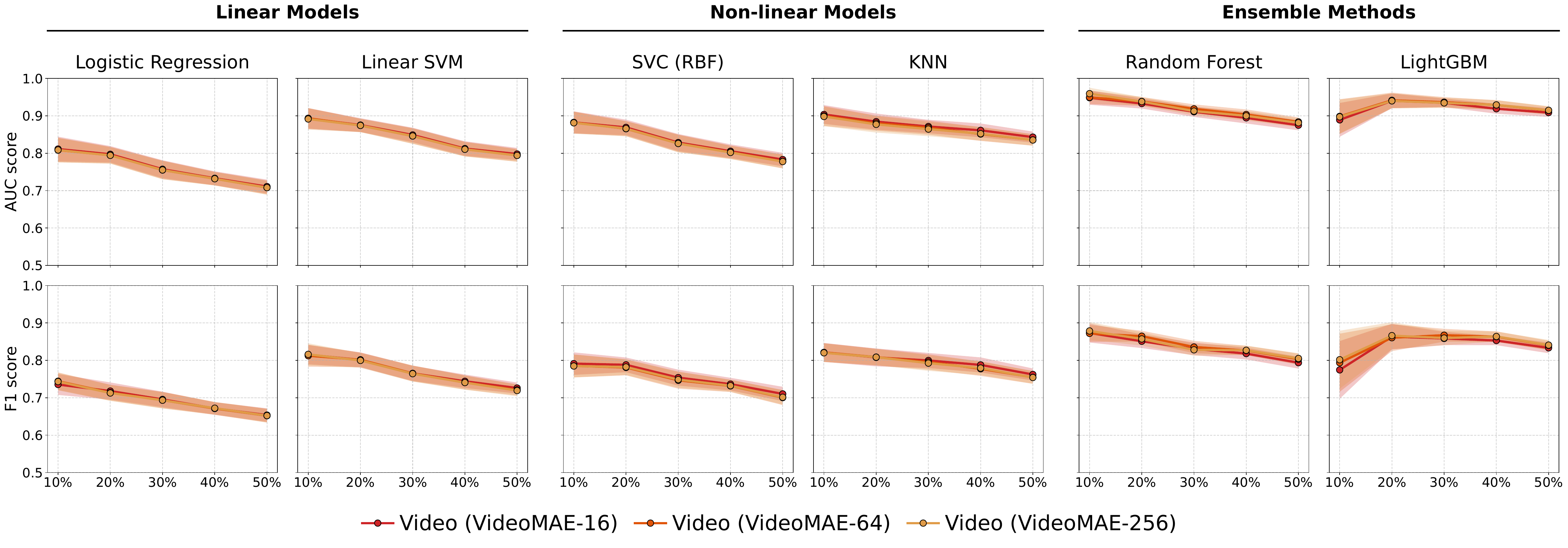}
    \caption{Binary classification performance (AUC, top; F1, bottom) for VideoMAE-v2 under three temporal sampling densities (16, 64, and 256 frames per clip), across six classifiers and five quantile thresholds. The three configurations produce nearly indistinguishable results, confirming that frame density does not drive performance differences between video and image representations.}
    \label{fig:videomae-frames}
\end{figure}

To assess whether the performance of spatiotemporal video features depends on the number of sampled frames, we evaluated VideoMAE-v2 under three temporal sampling densities: 16, 64, and 256 frames per clip. Figure~\ref{fig:videomae-frames} reports AUC and F1 scores across all six classifiers and five quantile thresholds for each configuration.

As shown, the three sampling densities yield nearly identical performance profiles across all classifier--threshold combinations, indicating that the observed advantages of video features over temporally compressed representations are not attributable to frame density. The marginal gains from increasing the frame count come at a substantially higher computational cost: relative to the 16-frame configuration, the 256-frame configuration requires approximately $16\times$ more frames to be processed per clip, with negligible performance improvement. We therefore report VideoMAE-16 as the primary video representation in the main text.



\subsection{Segment-level Analysis}\label{ssec:app-segment-frames}

To examine whether alignment varies across the engagement distribution, we divided clips within each video into three segments by heat value: low (bottom third), mid (middle third), and high (top third), and computed Spearman $\rho$ separately for each segment. Table~\ref{tab:spearman-segments} reports results for the primary representations.

Across all representations, alignment is strongest in the high segment and weakest in the mid segment, suggesting that engagement signals are most reliably encoded at the extremes of the distribution. This pattern holds across representational forms, indicating that segment-level difficulty is a property of the engagement signal rather than any particular representation.

\begin{table}[htbp]
    \centering
    \small
    \caption{Segment-level Spearman $\rho$ for primary representations. Low, mid, and high refer to thirds of the within-video heat value distribution.}
        \label{tab:spearman-segments}
    \begin{tabular}{lccc}
    \toprule
        \textbf{Representation} & \textbf{Low Score} & \textbf{Mid Score} & \textbf{High Score} \\
        \midrule
        Video (VideoMAE-256)            & 0.355 & 0.279 & 0.611 \\
        Video (VideoMAE-64)             & 0.334 & 0.291 & 0.590 \\
        Video (VideoMAE-16)             & 0.335 & 0.250 & 0.547 \\
        TAI (MAE ViT-L)                 & 0.209 & 0.197 & 0.464 \\
        Mid-point frame (MAE ViT-L)     & 0.163 & 0.190 & 0.396 \\
        Text (EVA-CLIP)                 & 0.202 & 0.179 & 0.343 \\
        Audio (VGGish)                  & 0.058 & 0.089 & 0.134 \\
        \bottomrule
    \end{tabular}
\end{table}
\section{Representational Divergence Analysis}\label{sec:app-gap-analysis}

\subsection{Prediction Correctness Cross-tabulation}\label{ssec:app-score-cross}

Table~\ref{tab:score-cross} reports the full cross-tabulation of prediction correctness categories by score level for the gap analysis reported in the main text.

\begin{table}[htbp]
    \centering
    \scriptsize
    \setlength{\tabcolsep}{6pt}
    \renewcommand{\arraystretch}{1.12}
    \caption{Cross-tabulation of correctness categories by score level. High- and low-score clips are defined based on replay heat quantiles within each video.}
    \label{tab:score-cross}
    \begin{tabular}{
        l
        S[table-format=4.0, round-mode=places, round-precision=0]
        S[table-format=4.0, round-mode=places, round-precision=0]
        S[table-format=4.0, round-mode=places, round-precision=0]
    }
    \toprule
    \textbf{Category} & \textbf{High Score} & \textbf{Low Score} & \textbf{Total} \\
    \midrule
    video ($\cmark$) \& TAI ($\cmark$)                    & 1287 & 1304  & 2591 \\
    video ($\xmark$) \& TAI ($\cmark$) -- $\mathcal{G}_A$ & 222  & 241   & 463  \\
    video ($\cmark$) \& TAI ($\xmark$) -- $\mathcal{G}_B$ & 102  & 80    & 182  \\
    video ($\xmark$) \& TAI ($\xmark$)                    & 95   & 100   & 195  \\
    \midrule
    \textbf{Total}                                          & 1706 & 1725  & 3431 \\
    \bottomrule
    \end{tabular}
\end{table}

\subsection{Semantic Segmentation Difference between Group A \& B}\label{ssec:app-segmentation}

\begin{figure}[htbp]
    \centering
    \includegraphics[width=1\columnwidth]{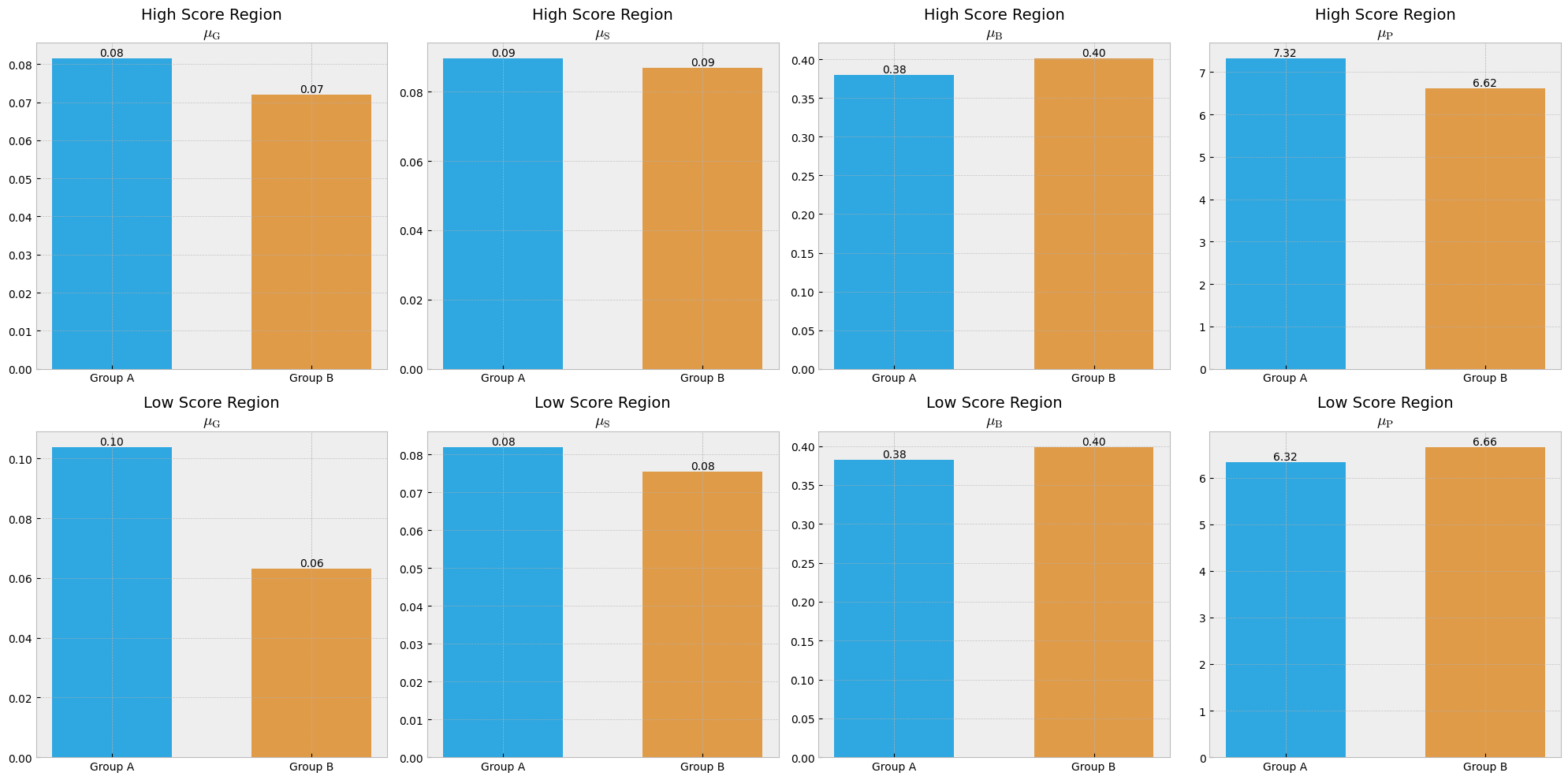}
    \caption{
        Clip-level mean proportions ($\mu$) of greenery, sky, building, and person count for $\mathcal{G}_A$ and $\mathcal{G}_B$, shown separately for high-score (top row) and low-score (bottom row) clips. Blue bars indicate $\mathcal{G}_A$ (TAI correct, video incorrect); orange bars indicate $\mathcal{G}_B$ (video correct, TAI incorrect).
    }
    \label{fig:segmentation-comp-mean}
\end{figure}

\begin{figure}[htbp]
    \centering
    \includegraphics[width=1\columnwidth]{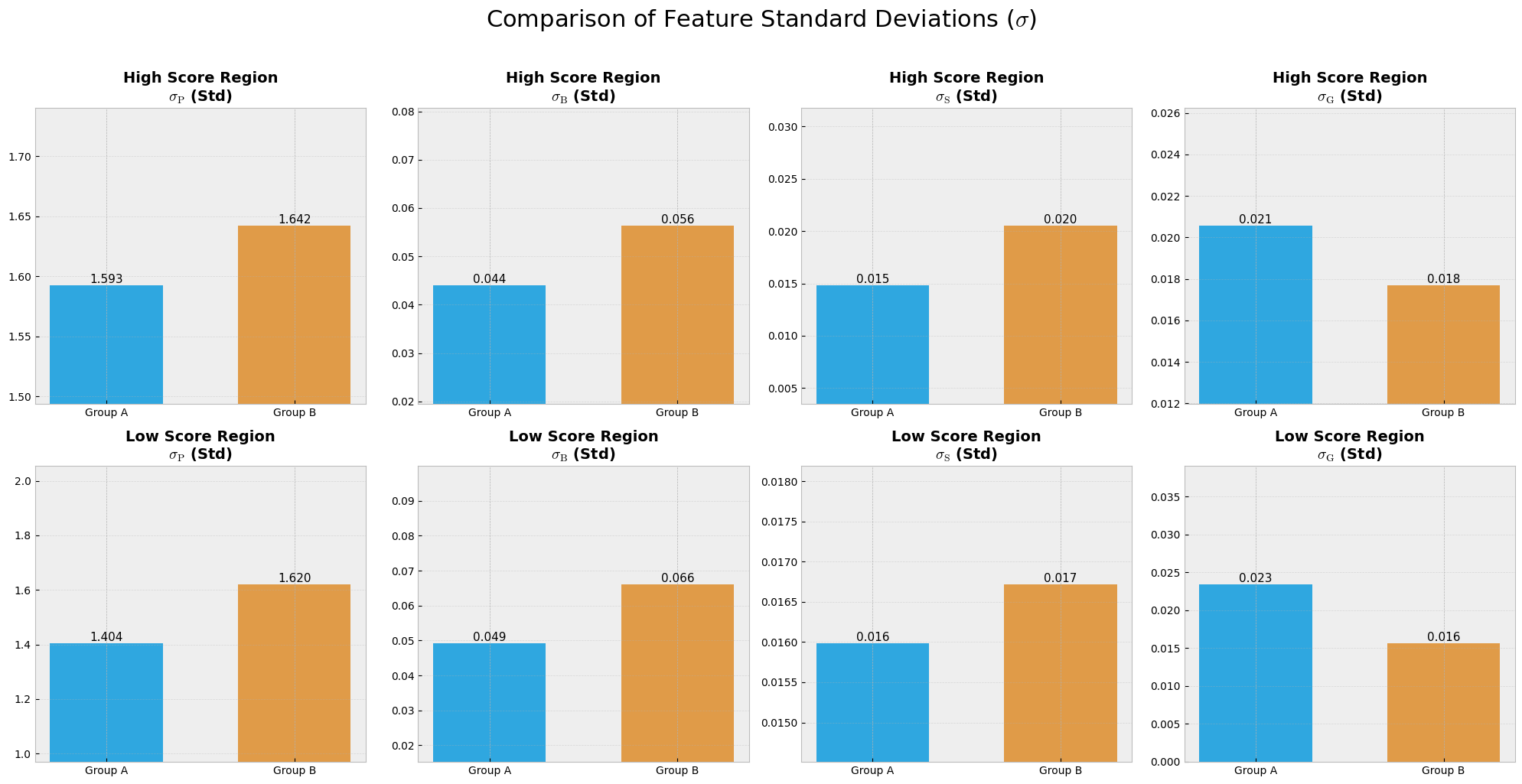}
    \caption{
        Within-clip standard deviations ($\sigma$) of greenery, sky, building, and person count for $\mathcal{G}_A$ and $\mathcal{G}_B$, shown separately for high-score (top row) and low-score (bottom row) clips. Higher $\sigma$ indicates greater frame-to-frame variability in scene composition within a clip. Color coding follows Fig.~\ref{fig:segmentation-comp-mean}.
    }
    \label{fig:segmentation-comp-std}
\end{figure}

Semantic descriptors are estimated at the clip level by applying a semantic segmentation model to each frame, yielding pixel-level proportions of buildings, sky, greenery, and persons. For each clip, frame-level estimates are averaged to yield a mean ($\mu$) reflecting overall scene composition, and the within-clip standard deviation across frames ($\sigma$) is retained as a measure of temporal variability. Both quantities are compared between $\mathcal{G}_A$ (blue) and $\mathcal{G}_B$ (orange) separately for high- and low-score clips (Figs.~\ref{fig:segmentation-comp-mean} and~\ref{fig:segmentation-comp-std}).

The mean descriptors reveal a consistent compositional difference between the two groups. $\mathcal{G}_A$ clips exhibit higher greenery coverage than $\mathcal{G}_B$ across both score regions (high-score: 0.08 vs.\ 0.07; low-score: 0.10 vs.\ 0.06), with the gap more pronounced in low-score clips. Building proportions show the opposite pattern, with $\mathcal{G}_B$ slightly but consistently higher (0.40 vs.\ 0.38 in both regions). Sky coverage is nearly identical across groups. Person count presents a score-contingent reversal: in high-score clips, $\mathcal{G}_A$ has a higher mean person density (7.32 vs.\ 6.62), whereas in low-score clips the pattern inverts (6.32 vs.\ 6.66 in favor of $\mathcal{G}_B$).

The standard deviation descriptors indicate that $\mathcal{G}_B$ clips are substantially more dynamic across all structural elements. Building proportion variance is markedly higher in $\mathcal{G}_B$ in both score regions (high: 0.056 vs.\ 0.044; low: 0.066 vs.\ 0.049), as is sky variance (high: 0.020 vs.\ 0.015; low: 0.017 vs.\ 0.016) and person count variance (high: 1.642 vs.\ 1.593; low: 1.620 vs.\ 1.404). Greenery variance is the sole exception, remaining higher in $\mathcal{G}_A$ (high: 0.021 vs.\ 0.018; low: 0.023 vs.\ 0.016). Together, these patterns suggest that $\mathcal{G}_B$ clips involve greater frame-to-frame changes in scene composition, consistent with dynamic street contexts, where video representations provide a necessary advantage over static visual summaries.

\subsection{Representative Clip Examples from \texorpdfstring{$\mathcal{G}_A$}{GA} and \texorpdfstring{$\mathcal{G}_B$}{GB}}\label{ssec:app-clip-examples}

The following figures provide representative visual examples from each group, illustrating the scene types associated with $\mathcal{G}_A$ (TAI correct, video incorrect) and $\mathcal{G}_B$ (video correct, TAI incorrect). Within each figure, the left column shows high-score clips and the right column shows low-score clips.

\begin{figure}[htbp]
    \centering
    \includegraphics[width=0.92\linewidth,height=0.78\textheight,keepaspectratio]{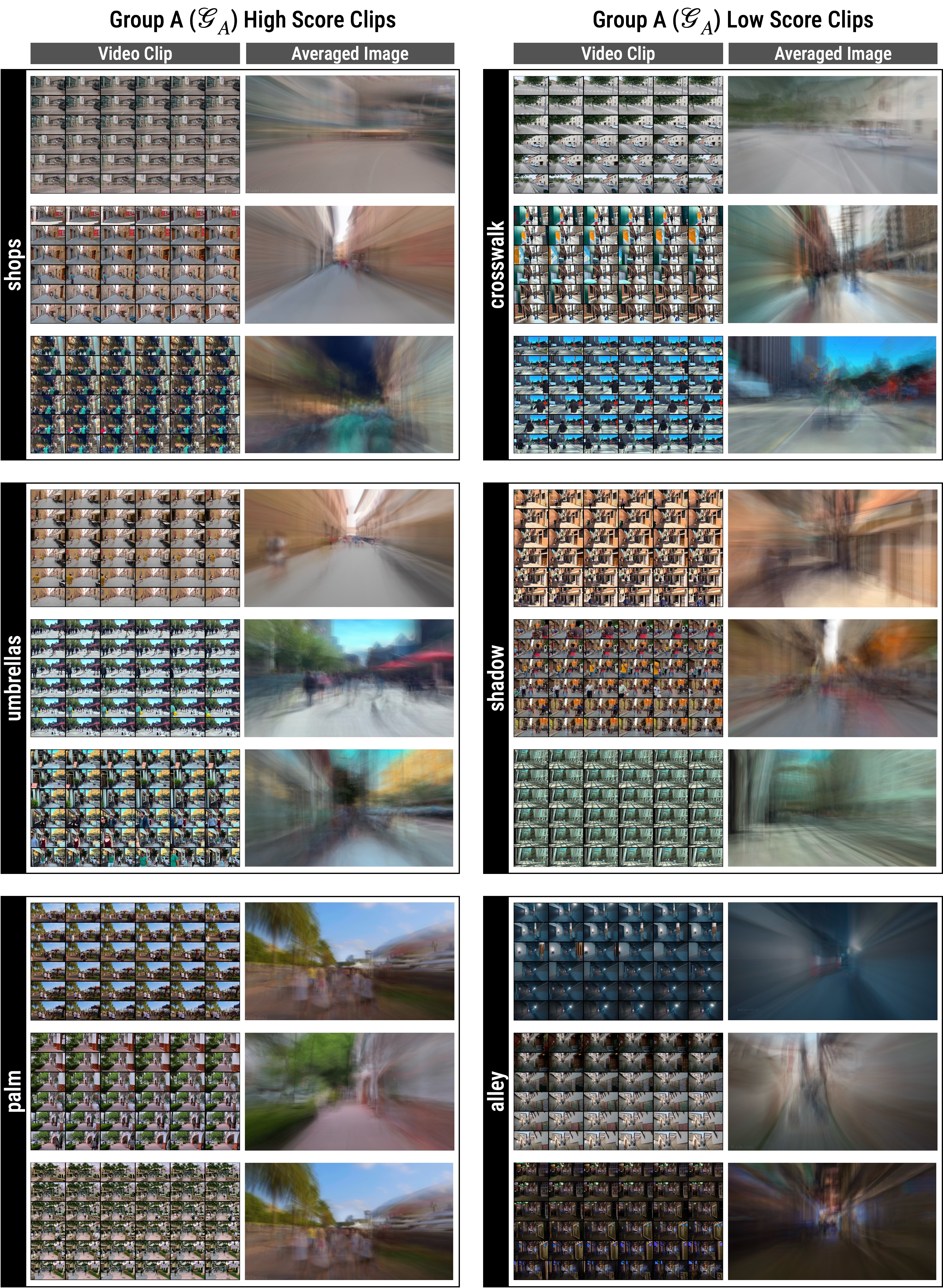}
    \caption{Representative clips from $\mathcal{G}_A$ (TAI correct, video incorrect). Left column: high-score clips; right column: low-score clips. Across both score levels, scenes are characterized by stable spatial composition, large homogeneous regions, and weak temporal variation.}
    \label{fig:clip-image-group-a}
\end{figure}

\begin{figure}[htbp]
    \centering
    \includegraphics[width=0.92\linewidth,height=0.78\textheight,keepaspectratio]{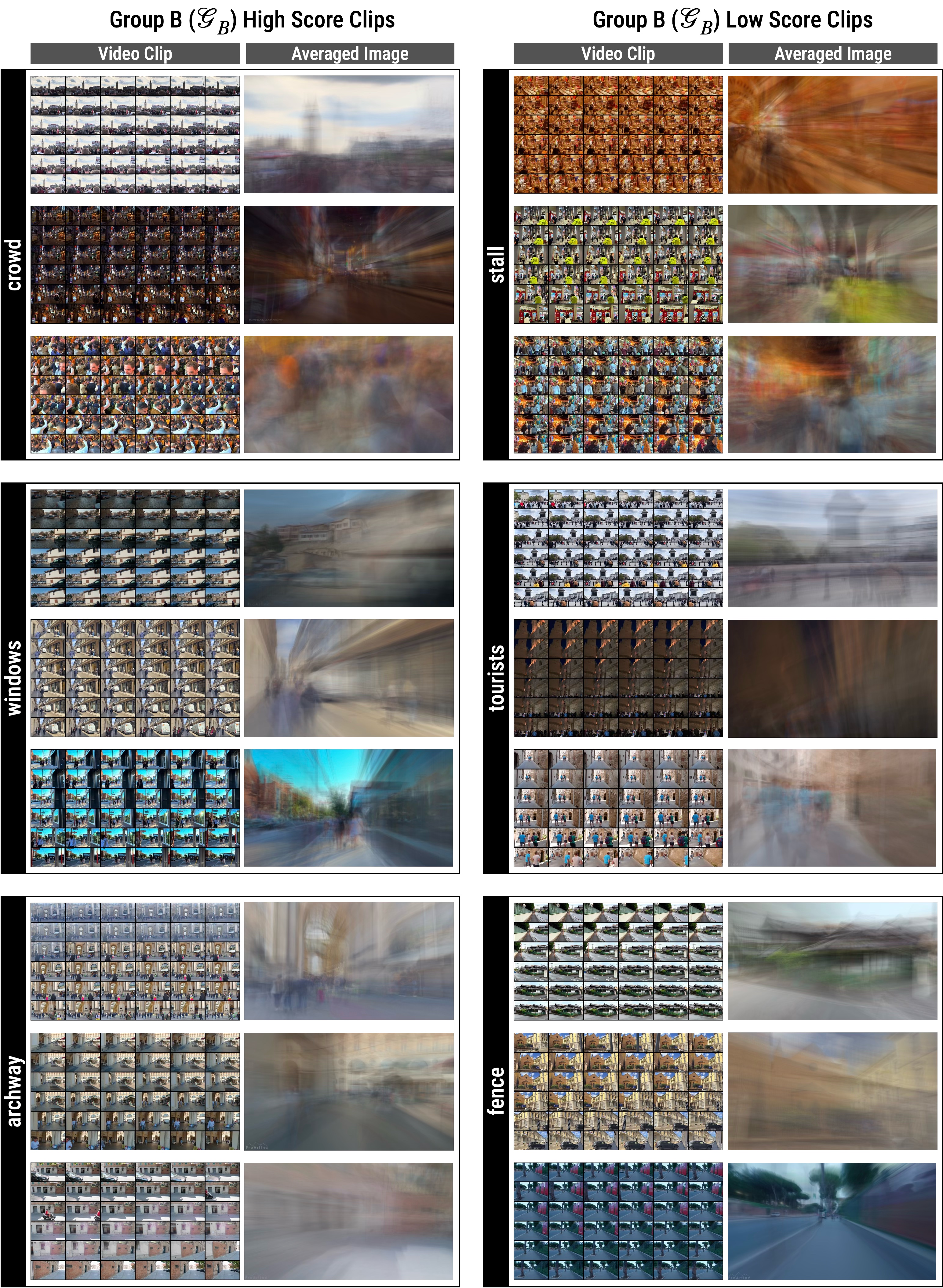}
    \caption{Representative clips from $\mathcal{G}_B$ (video correct, TAI incorrect). Left column: high-score clips; right column: low-score clips. Across both score levels, scenes contain dynamic content---dense pedestrian activity, moving vehicles, and rapidly changing viewpoints---that the TAI collapses into indistinct composites, consistently favoring video representations regardless of engagement level.}
    \label{fig:clip-image-group-b}
\end{figure}

\section{Human Validation Details}\label{sec:app-amt}

\begin{figure}[htbp]
    \centering
    \begin{subfigure}[t]{0.48\linewidth}
        \centering
        \includegraphics[width=\linewidth]{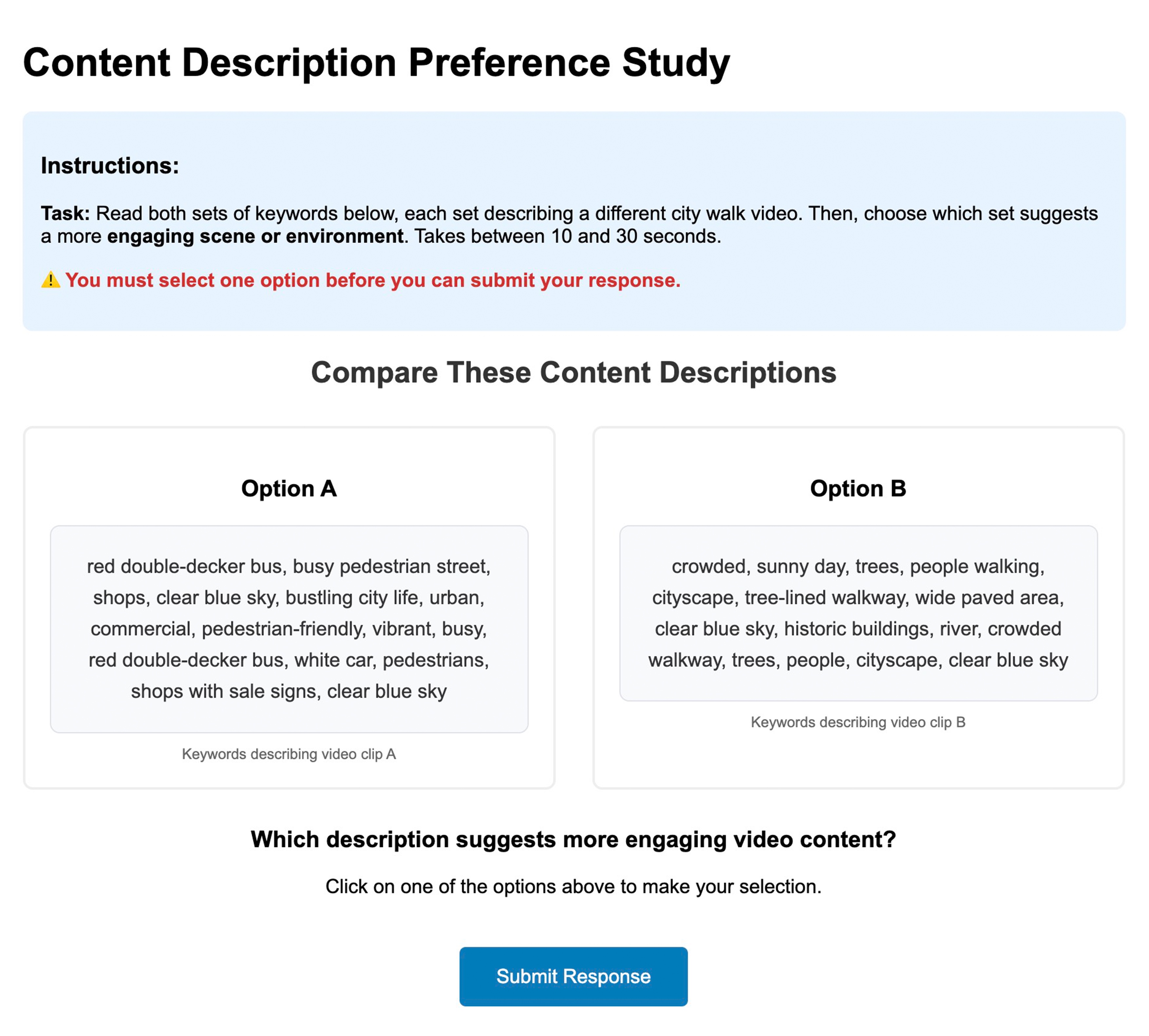}
        \caption{Keyword-based task}
    \end{subfigure}
    \hfill
    \begin{subfigure}[t]{0.48\linewidth}
        \centering
        \includegraphics[width=\linewidth]{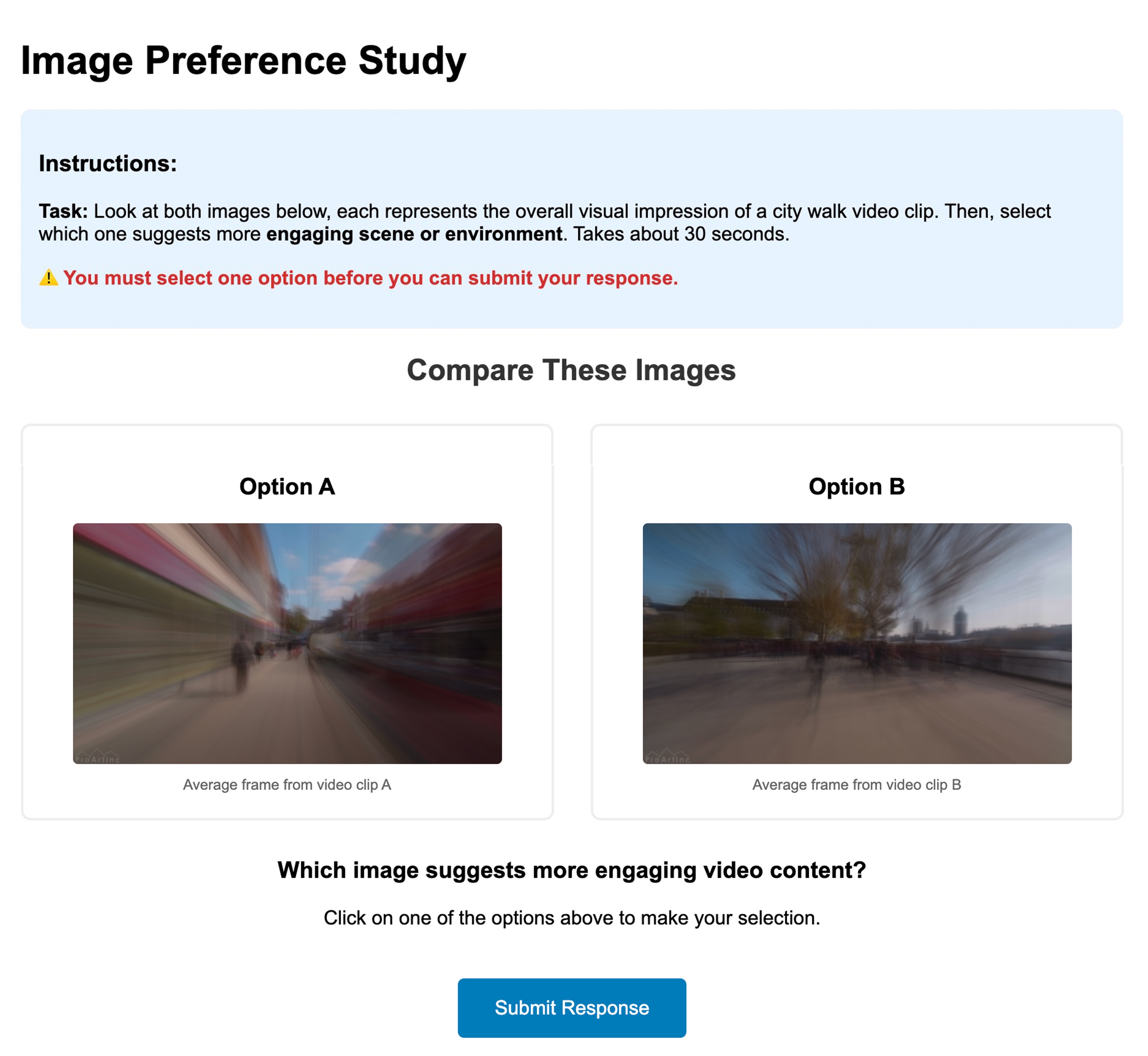}
        \caption{TAI-based task}
    \end{subfigure}
    \vspace{0.6em}
    \begin{subfigure}[t]{0.48\linewidth}
        \centering
        \includegraphics[width=\linewidth]{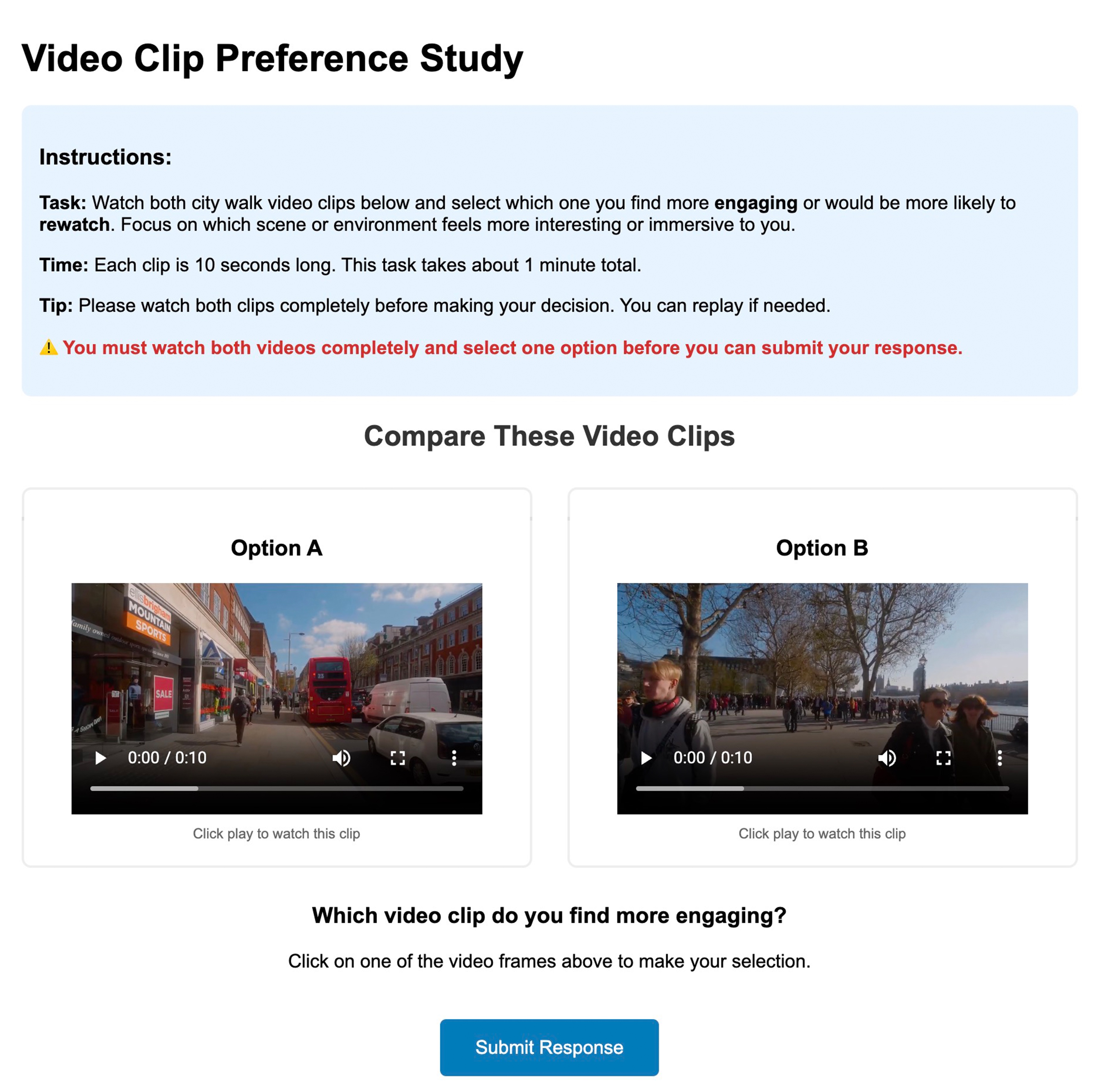}
        \caption{Video-based task}
    \end{subfigure}
    \caption{AMT interfaces for the three preference tasks. Each task presents a pair of moments (A/B) drawn from the same source video and applies a consistent decision rule across representational conditions.}
    \label{fig:amt-survey}
\end{figure}

Full details of the human annotation protocol, including worker eligibility criteria, the three task designs, and quality control procedures, are provided in the Methods section of the main text. Example interfaces for all three tasks are shown in Fig.~\ref{fig:amt-survey}.

\begin{figure}[htbp]
    \centering
    \includegraphics[width=1\columnwidth]{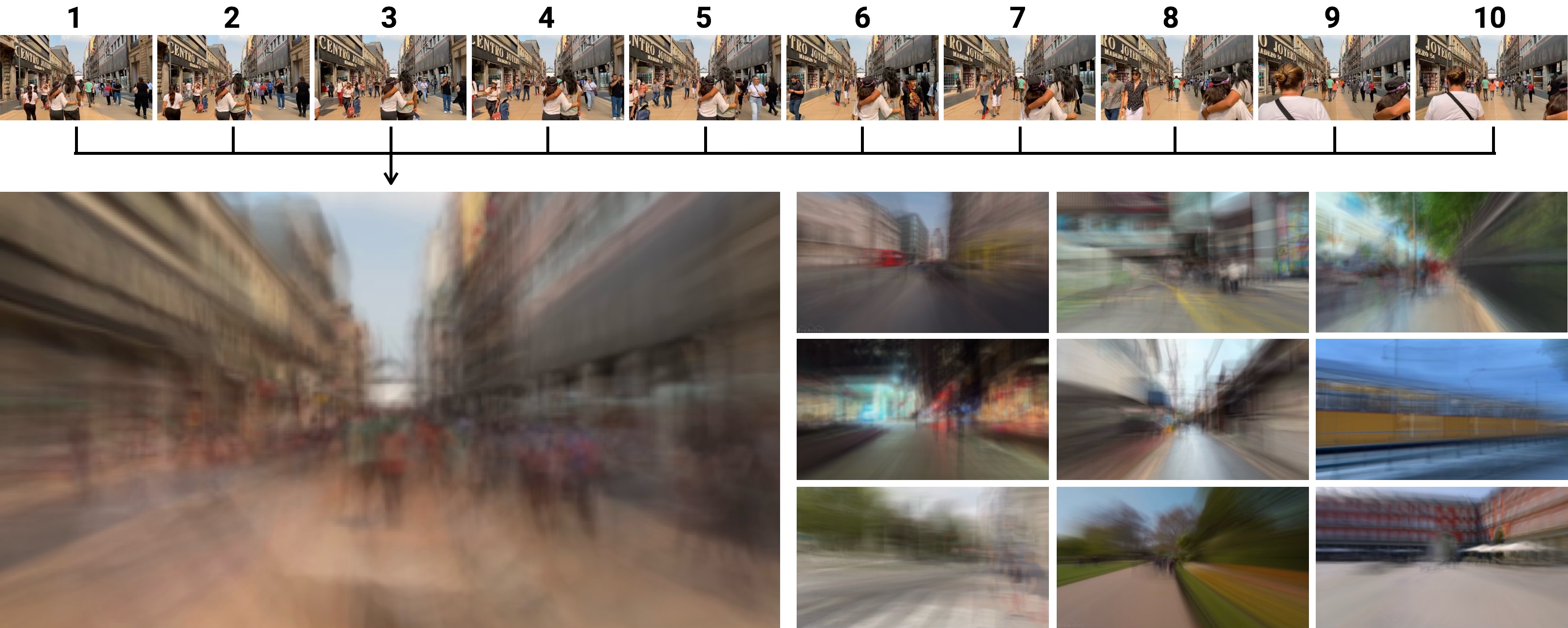}
    \caption{Illustration of the temporally averaged image (TAI) representation. Top: uniformly sampled frames (1~Hz) from a single 10-second clip. Bottom left: the resulting pixel-wise average, which suppresses transient elements while preserving stable spatial structure. Bottom right: example TAIs from other clips in the corpus.}
    \label{fig:supp-averaged-image}
\end{figure}

\end{document}